\documentclass[10pt,twocolumn,letterpaper]{article}
\PassOptionsToPackage{table}{xcolor}

\usepackage[pagenumbers]{cvpr}
\usepackage[utf8]{inputenc}
\usepackage[T1]{fontenc}
\usepackage{microtype}
\usepackage{graphicx}
\usepackage{booktabs}
\usepackage{multirow}
\usepackage{makecell}
\usepackage{array}
\usepackage{tabularx}
\usepackage{adjustbox}
\usepackage{threeparttable}
\usepackage{wrapfig}
\usepackage{subcaption}
\usepackage{float}
\usepackage{amsmath,amsfonts,amssymb}
\usepackage{mathtools}
\usepackage{amsthm}
\usepackage{bm}
\usepackage{bbm}
\usepackage{xspace}
\usepackage{pifont}
\usepackage{ragged2e}
\usepackage[most]{tcolorbox}
\tcbuselibrary{breakable,skins,theorems}
\usepackage[linesnumbered,ruled,vlined]{algorithm2e}
\usepackage{soul}
\usepackage{tikz}
\usetikzlibrary{positioning,arrows.meta,fit,shadows,calc}

\definecolor{LLMain}{HTML}{1F4E79}
\definecolor{LLAccent}{HTML}{4B9CD3}
\definecolor{LLSoft}{HTML}{EEF6FC}
\definecolor{LLSoftTwo}{HTML}{F6FAFD}
\definecolor{LLDark}{HTML}{16324F}
\definecolor{LLGood}{RGB}{34,139,75}
\definecolor{LLBad}{RGB}{195,65,55}
\definecolor{LLWarn}{RGB}{220,140,35}
\definecolor{LLGray}{gray}{0.42}
\definecolor{LLHeader}{RGB}{240,242,245}
\definecolor{LookOne}{HTML}{DCEEFF}
\definecolor{LookTwo}{HTML}{A9D1F6}
\definecolor{LookThree}{HTML}{5C9ED6}
\definecolor{ThinkColor}{HTML}{7B6FD6}
\definecolor{RouteColor}{HTML}{F5A45D}
\definecolor{LLDeltaUp}{HTML}{008F59}
\definecolor{LLDeltaDown}{HTML}{F02035}

\newcommand{\cmark}{\textcolor{LLGood}{\ding{51}}}
\newcommand{\xmark}{\textcolor{LLBad}{\ding{55}}}

\newcommand{\accgood}[1]{$_{\color{LLDeltaUp}\uparrow #1}$}
\newcommand{\accbad}[1]{$_{\color{LLDeltaDown}\downarrow #1}$}

\definecolor{toponecell}{RGB}{211,234,214}
\definecolor{toptwocell}{RGB}{232,241,204}
\definecolor{topthreecell}{RGB}{248,241,203}

\newcolumntype{I}{!{\vrule width 1pt}}
\makeatletter
\newcommand{\thickhline}{%
  \noalign{\ifnum0=`}\fi\hrule height 1pt
  \futurelet\reserved@a\@xhline
}
\makeatother

\newtcolorbox{insightbox}[1]{
  enhanced jigsaw, breakable,
  colback=LLMain!4!white, colframe=LLMain!82!black,
  boxrule=0.8pt, arc=4pt,
  left=7pt, right=7pt, top=6pt, bottom=4pt,
  before skip=8pt, after skip=8pt,
  fontupper=\small,
  title={#1}, colbacktitle=LLMain, coltitle=white,
  fonttitle=\scriptsize\bfseries,
  attach boxed title to top left={xshift=12pt,yshift=-2.6mm},
  boxed title style={colback=LLMain, colframe=LLMain, boxrule=0pt, arc=2pt,
    left=6pt, right=6pt, top=2pt, bottom=2pt}
}

\newtcolorbox{positionbox}[1][Positioning]{
  enhanced, breakable,
  colback=LLSoftTwo, colframe=LLDark,
  boxrule=0.6pt, arc=3pt,
  left=7pt, right=7pt, top=5pt, bottom=5pt,
  before skip=7pt, after skip=7pt,
  fontupper=\small,
  title={#1}, colbacktitle=LLDark, coltitle=white,
  fonttitle=\scriptsize\bfseries
}

\newtcolorbox{takeawaybox}[1][Takeaway]{
  enhanced, breakable,
  colback=LLAccent!6!white, colframe=LLAccent!70!black,
  boxrule=0.55pt, arc=3pt,
  left=6pt, right=6pt, top=5pt, bottom=5pt,
  before skip=6pt, after skip=6pt,
  fontupper=\small,
  title={#1}, colbacktitle=LLAccent!80!black, coltitle=white,
  fonttitle=\scriptsize\bfseries
}

\newtcolorbox{promptbox}[1][Prompt]{
  enhanced, breakable,
  colback=LLSoftTwo, colframe=LLMain!50!black,
  fontupper=\normalsize,
  boxrule=0.6pt, arc=3pt,
  left=8pt, right=8pt, top=5pt, bottom=6pt,
  title={\normalsize\bfseries #1}, fonttitle=\bfseries,
  before skip=8pt, after skip=8pt
}

\makeatletter
\newcommand{\tightparagraph}{\@startsection{paragraph}{4}{\z@}{0.8ex plus 0.2ex minus 0.1ex}{-1em}{\normalsize\bf}}
\newcommand{\baselineparagraph}{\@startsection{paragraph}{4}{\z@}{1.2ex plus 0.3ex minus 0.1ex}{-1em}{\normalsize\bf}}
\newcommand{\methodparagraph}{\@startsection{paragraph}{4}{\z@}{1.8ex}{-1em}{\normalsize\bf}}
\makeatother

\SetKwInput{Input}{Input}
\SetKwInput{Output}{Output}
\DontPrintSemicolon

\SetCommentSty{llcommfont}
\SetKwComment{Comment}{$\triangleright$ }{}

\theoremstyle{plain}

\theoremstyle{definition}

\theoremstyle{remark}

\AtBeginDocument{%
  \crefname{figure}{Fig.}{Figs.}%
  \Crefname{figure}{Figure}{Figures}%
  \crefname{algorithm}{Alg.}{Algs.}%
  \Crefname{algorithm}{Algorithm}{Algorithms}%
  \crefname{appendix}{Appendix}{Appendices}%
  \Crefname{appendix}{Appendix}{Appendices}%
  \crefname{proposition}{Prop.}{Props.}%
  \Crefname{proposition}{Proposition}{Propositions}%
}

\definecolor{linkblue}{rgb}{0.21,0.49,0.74}
\usepackage[pagebackref,breaklinks,colorlinks,allcolors=linkblue]{hyperref}

\title{Loop to Look Better: Learning Adaptive Depth for Question-Guided Perception in Vision-Language Models}

\author{
    Wencheng Ye$^{1,*}$,
    Anning Hu$^{2,*}$,
    Hongye Fang$^{1}$,
    Tianyi Wang$^{1}$,
    Junchi Yan$^{2,\dagger}$
    \\
    $^{1}$Tongji University
    \qquad
    $^{2}$Shanghai Jiao Tong University
    \\
    $^{*}$Equal contribution
    \qquad
    $^{\dagger}$Corresponding author
}
\date{}
\hypersetup{
  pdftitle={Loop to Look Better: Learning Adaptive Depth for Question-Guided Perception in Vision-Language Models},
  pdfauthor={},
  pdfsubject={Preprint},
  pdfkeywords={LoopLook, vision-language models, adaptive depth, recurrent perception}
}

\begin{document}
\maketitle

\begin{abstract}
Visual perception should be iterative and task-directed, with representations refined in response to the question and the need for further processing. While Vision-Language Models (VLMs) increasingly support extended reasoning, their visual tokens are still typically acquired in a single, task-agnostic encoder pass. This creates an evidence-acquisition bottleneck: models may reason longer without refining the visual evidence needed to answer the question. We propose \textbf{LoopLook}, a lightweight encoder-side recurrent perception framework that equips pretrained VLMs with adaptive visual depth. LoopLook recurrently revisits a middle-to-late window of the vision encoder, injecting question representations at each loop to guide visual refinement. A learned stopping mechanism adaptively determines the recurrent depth, while a token-wise \textbf{Mixture-of-Looks} (MoL) combines representations from different loops, allowing each visual token to selectively integrate information acquired across recurrent steps. Across four backbones and eight benchmarks, LoopLook consistently improves performance over the baseline, yielding gains of up to 8.6\% on MMStar and 8.1\% on HallusionBench.
% and 16.7\% on the OCR task of V*
Our analysis shows that LoopLook progressively refines visual attention and adaptively allocates depth according to task difficulty. It also better preserves task-relevant visual information under reduced input resolution, while increasing TTFT by only 5.1–6.5\%. By bringing recurrence from reasoning back into perception, LoopLook transforms visual perception from a one-shot glimpse into an adaptive process of recurrent refinement.  
% Code: \href{https://github.com/gooogleshanghai/looplook}{LoopLook}.

\end{abstract}
\vspace{-1em}
\section{Introduction}
\label{sec:intro}

Visual perception is inherently \textit{iterative} and \textit{task-directed}: an initial
observation provides a coarse understanding of a scene, while
subsequent processing refines this understanding according to
the task~\cite{PARR201792,LAMME2000571}. In human vision, this process involves revisiting ambiguous regions, inspecting small objects, and actively seeking relevant cues~\cite{yarbus2013eye,hayhoe2005eye}.
Drawing on this interplay between iterative refinement and task guidance, we ask a natural question for Vision-Language Models (VLMs): \textit{How can visual processing be precisely guided by the question and adaptively extended when needed?}

\begin{figure}[t]
    \centering
    \includegraphics[width=\columnwidth]{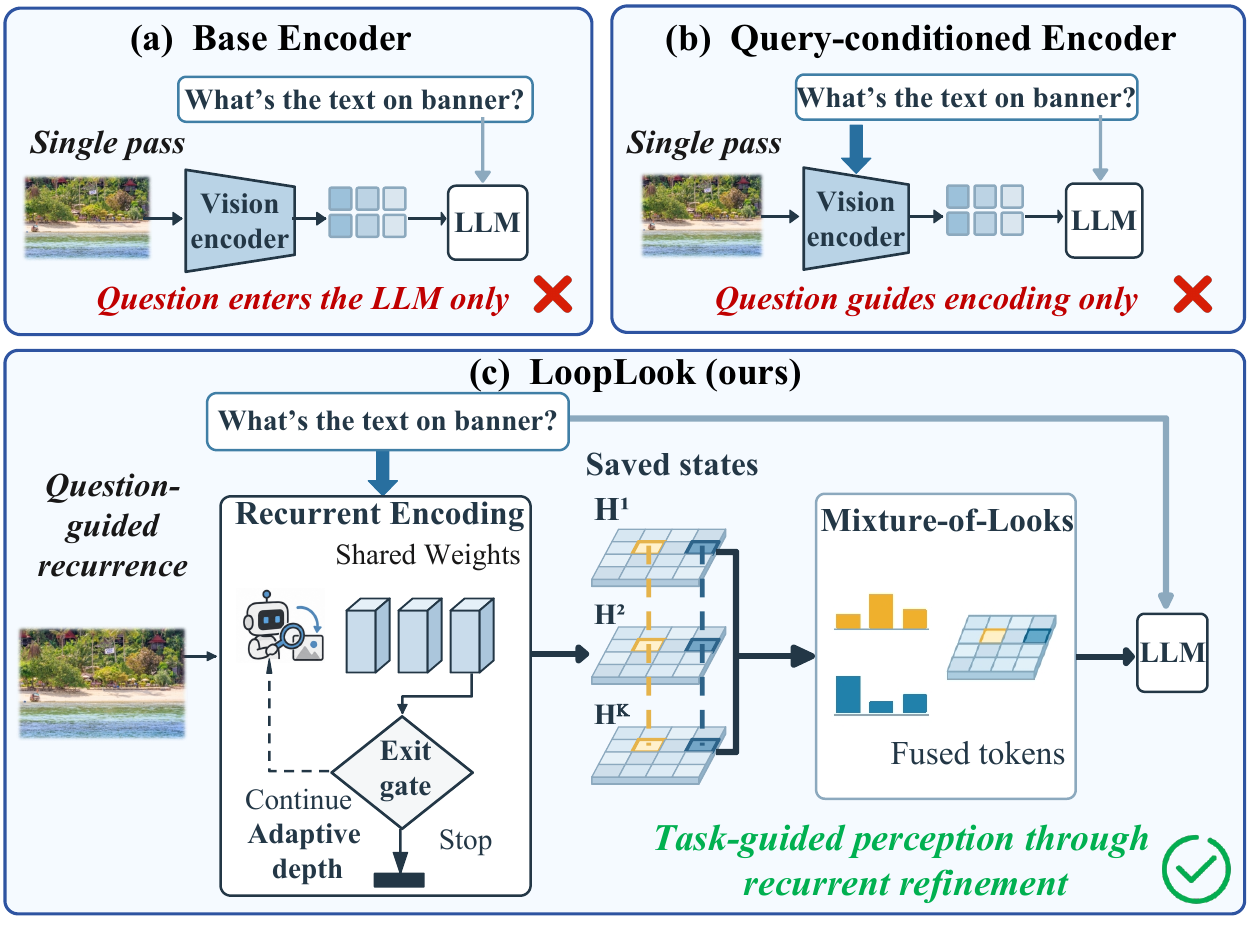}
    \caption{\textbf{Comparison of visual encoding paradigms.} (a) The base encoder processes inputs in a single pass. (b) Query-conditioned encoding incorporates the question into encoding but remains single-pass. (c) LoopLook enables task-guided refinement with adaptive computation depth.}
    \label{fig:overview}
\end{figure}
Recent advances have explored additional computation in VLMs, expanding the reasoning budget through explicit reasoning traces~\cite{pmlr-v306-hu26ba,wei2022chain,wang2022self,shao2024visual}, recursive refinement~\cite{wang2026vl,cheng2025vision,he2025self}, and latent recurrent computation~\cite{pham2025multimodal,jeon2026vision,li2026latent}. These methods give models more opportunities to interpret multimodal context after the image has been encoded. Still, the visual evidence itself is acquired in a single, task-agnostic pass: the image is converted into visual tokens before the model can determine which regions, details, or relations are required by the question~\cite{dai2023instructblip,prisadnikov2025vision}. This creates an \textbf{evidence-acquisition bottleneck} between \textit{perception} and \textit{cognition}: a VLM may reason longer, yet still lack task-critical visual evidence that was inadequately captured during the initial encoding, potentially leading to insufficient visual grounding or overreliance on linguistic priors~\cite{tong2024eyes,pmlr-v306-wu26f,pmlr-v235-lin24c,ye2026ebb}. Such a bottleneck is particularly consequential for tasks requiring fine-grained perception or image-dependent reasoning, where additional reasoning cannot compensate for underprocessed visual information~\cite{chen2026babyvision,wu2024v}.

A direct way to address this limitation is to improve visual representations and their interface to the language model. Existing methods have pursued this goal by increasing input resolution~\cite{liu2024llavanext,li2024monkey,li2025mini}, aggregating features from multiple encoder layers~\cite{yao2024dense,meng2024deepstack}, or using the instruction to fuse task-relevant visual layers~\cite{lin2026text,wu2024v,kim2026mol}. These advances show that VLM performance depends critically on how visual evidence is extracted, selected, and exposed. However, better selection and delivery of visual features do not, by themselves, enable the encoder to iteratively refine visual representations according to each question's perceptual demands.

Adaptive recurrent computation provides exactly the mechanism needed to bring this missing flexibility into visual perception. Looped Transformers repeatedly apply a shared block to the hidden state, increasing effective computation depth without introducing separate parameters~\cite{dehghani2018universal,geiping2026scaling}. Such recurrence has proven especially effective for language modeling~\cite{wang2026smelt,zhu2025scaling}, with adaptive variants further adjust computation through learned halting~\cite{graves2016adaptive,dehghani2018universal}, budget-conditioned trajectories~\cite{jeddi2026loopformer}, or token-level routing~\cite{bae2026mixture,you2026improvingtesttimescalingadaptive}. In vision and vision-language models, early explorations of looped architectures suggest that repeated computation can likewise refine visual representations and improve abstract or multimodal reasoning~\cite{qian2026loopvl,shu2026loopvit,burzio2026d}. Together, these findings point to the potential of revisiting encoder layers to iteratively refine visual representations.

% Realizing this potential requires coordinating how recurrent computation is performed and how its outputs are used. The updates should be guided by the question and evolving visual state, while the number of recurrent steps should adapt to each input. Moreover, different tokens may benefit from different mixtures of recurrent states: tokens covering a queried object and its surrounding context may rely on earlier and later representations to different extents. 

To realize this potential, we propose \textbf{LoopLook}, a lightweight encoder-side recurrent perception framework that
  equips pretrained VLMs with adaptive visual depth.
  LoopLook reuses a middle-to-late encoder window to iteratively update the visual state
  with shared weights (\Cref{fig:overview}). 
  At each loop, the instruction representation, current visual
  summary, and loop-step embedding jointly generate layer-wise
  affine modulation, adapting the shared transformations to the
  task and the evolving state.
   A learned exit gate evaluates the visual state after each update
  to determine whether another loop should be executed.
  Once execution stops, LoopLook constructs the final visual
  representation from the accumulated states rather than relying
  solely on the last update.
  To accommodate regional differences in how features evolve
  across loops, token-wise \textbf{Mixture-of-Looks} (MoL) learns
  a separate weighting over these states for each visual token.
  By combining LoRA with trainable modulation and keeping the language model frozen,
  LoopLook enables task-guided iterative refinement through
  lightweight adaptation.

We evaluate LoopLook on four pretrained VLM backbones across eight downstream benchmarks. LoopLook consistently improves average performance across all backbones, with gains of up to $8.6$\% on MMStar and $8.1$\% on HallusionBench over the base model. Across recurrent loops, LoopLook progressively refines visual attention toward task-relevant evidence, while its token-wise Mixture-of-Looks selectively integrates representations from different loops. Under reduced input resolution, LoopLook mitigates visual degradation by allocating greater recurrent depth and better preserving task-relevant information. Analysis of exit depths further shows that LoopLook assigns more loops to fine-grained and hallucination-prone tasks. This adaptive perception incurs only modest overhead, increasing time to first token (TTFT) by 5.1–6.5\% while leaving autoregressive decoding essentially unchanged. Our contributions are summarized as follows:

\begin{itemize}
    \item \textbf{\textit{Bottleneck Identification.}} We formulate the one-shot evidence-acquisition bottleneck in VLMs, where cognition can receive additional computation while visual evidence remains fixed after a single, task-agnostic encoder pass.
    \item \textbf{\textit{Recurrent Framework.}} We propose LoopLook, a recurrent perception framework that reuses a middle-to-late visual encoder window with task-conditioned modulation, adaptive exit, and Mixture-of-Looks.
    \item \textbf{\textit{Comprehensive Evaluation.}} We evaluate LoopLook on four VLM backbones and eight benchmarks, demonstrating consistent gains, robustness to reduced resolution, learned depth allocation, and modest runtime overhead.
\end{itemize}
\section{Related Work}
\label{sec:related}
\paragraph{Vision Encoders in Multimodal Models.}
VLMs typically integrate pretrained visual encoders with language models through a modality-alignment interface that maps visual features into the language model's representation space~\cite{liu2024llavanext,vasu2025fastvlm,dai2023instructblip,bai2025qwen25vltechnicalreport}, with CLIP and SigLIP exemplifying visual encoders pretrained to align image representations with natural language supervision~\cite{radford2021clip,zhai2023sigmoid}. To understand the resulting representations, studies further examine cross-modal convergence~\cite{huh2024platonic}, develop tools for measuring representation similarity~\cite{kornblith2019similarity}, and analyze the structure and utility of intermediate visual features~\cite{raghu2021vision,caron2021emerging,prisadnikov2025vision,bolya2026perception}. However, extracting the visual evidence needed for grounded perception and reasoning remains challenging for VLMs~\cite{tong2024eyes,prisadnikov2025vision,chen2024we}. Two complementary lines of work address this challenge. One incorporates language conditions into feature extraction, ranging from early feature-wise modulation~\cite{de2017modulating,perez2018film} to question-aware visual encoders~\cite{kim2026tabflash,ganz2024question,liu2026igvlm,mao2026language,abramovich2024visfocus}. The other improves how visual features are selected and passed to the language model through visual token packing, multi-layer aggregation, layer fusion, and routing~\cite{li2025tokenpacker,yao2024dense,lin2026text,kim2026mol,meng2024deepstack}.
\begin{figure*}[t]
    \centering
    \includegraphics[width=\textwidth]{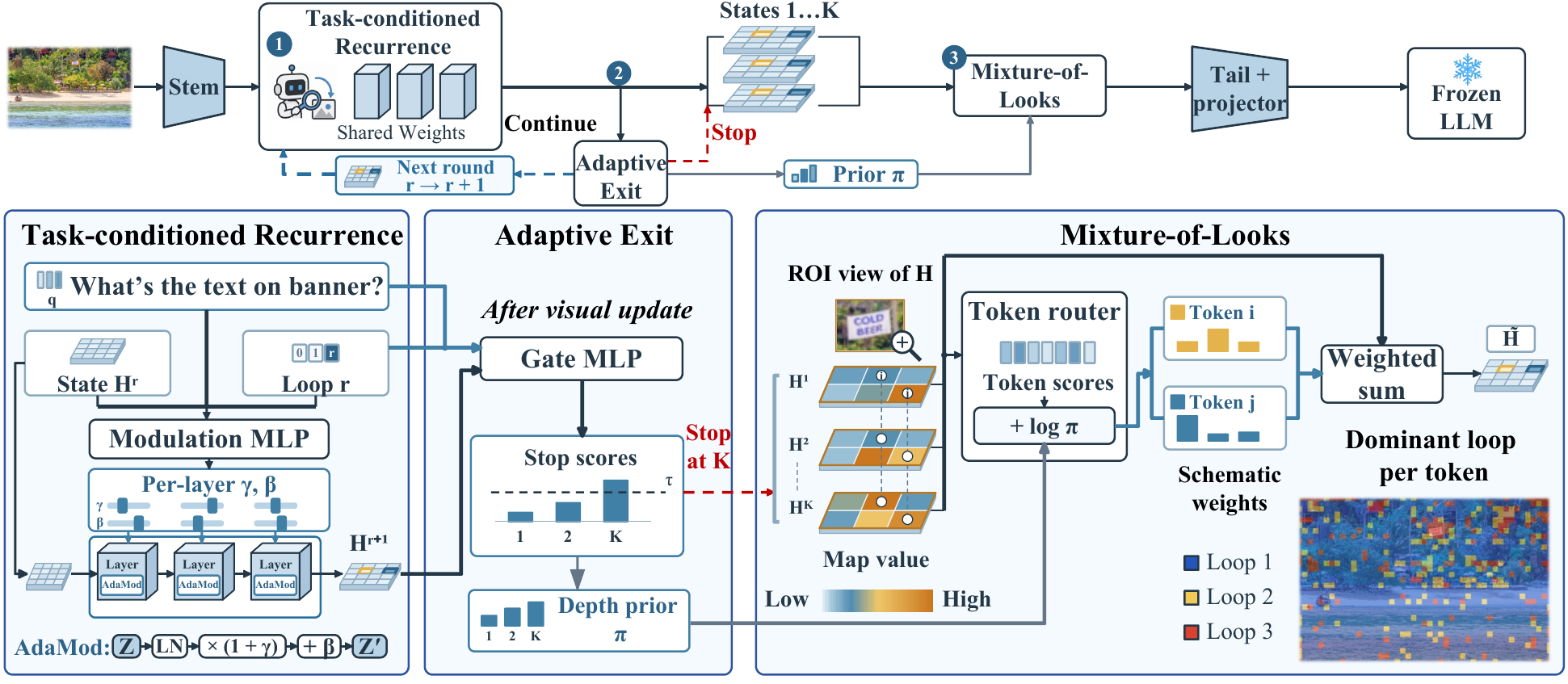}
    \caption{\textbf{Overview of LoopLook framework.} 
}
    \label{fig:framework}
\end{figure*}
\vspace{-0.2em}
\paragraph{Adaptive Computation in Recurrent Networks.}
ACT introduced learned halting for recurrent
  computation~\cite{graves2016adaptive}, while Universal Transformers
  combined weight-shared updates with adaptive processing
  depth~\cite{dehghani2018universal}.
  In language models, recurrent-depth architectures explore latent
  reasoning and compute-efficient scaling through repeated
  updates~\cite{zhu2025scaling,geiping2026scaling,wang2026smelt},
  while other approaches adapt pretrained models for shared-depth
  execution~\cite{bae2025relaxed,mcleish2025teaching}.
  Beyond increasing effective depth, recent work studies how to
  control this computation through budget-conditioned
  execution~\cite{jeddi2026loopformer}, token-specific recursion
  depths~\cite{bae2026mixture}, and learned stopping for test-time
  scaling~\cite{you2026improvingtesttimescalingadaptive}.
  In vision, recurrent processing has been explored for convolutional
  feature refinement~\cite{liang2015recurrent,linsley2018learning}, image classification~\cite{lee2026soft}, 
  abstract visual reasoning~\cite{shu2026loopvit}, and multi-view
  reconstruction~\cite{burzio2026d}.
  Within recursive vision architectures, block-recurrent designs re-execute a shared transformer block across depths~\cite{jacobs2026block}. Building on this view, reViT programs depth-specific experts into the shared block~\cite{bulat2026one}.
  Extending recurrence to multimodal modeling, LoopVL combines loops to iteratively update a
  unified vision--language state~\cite{qian2026loopvl}.

\paragraph{Positioning of Our Work.} LoopLook connects these two research directions by making visual feature extraction both question-conditioned and recurrent. While most recurrent approaches train architectures from scratch, it requires only lightweight adaptation, preserving the pretrained visual representations while enabling iterative refinement.

% Developed concurrently and independently, AdaLoop~\cite{han2026adaloop} explores question-conditioned recurrence in vision encoders but requires decoding at each depth for answer-consistency halting. LoopLook combines learned encoder-side stopping
%   with token-wise fusion, enabling recurrent refinement with modest
%   prefill overhead and no repeated answer decoding.
\section{Method}
\label{sec:method}

LoopLook refines visual representations through task-conditioned recurrence within a pretrained vision encoder before answer generation (\Cref{fig:framework}). Its design combines a shared recurrent window, loop-dependent feature modulation, and a Mixture-of-Looks module for adaptive stopping and token-wise state fusion. In this section, we first introduce the visual encoding formulation, then describe these components and the training objective.

\subsection{Preliminary}
\begin{figure}[t]
    \centering
    \begin{subfigure}[t]{0.49\linewidth}
        \centering
        \includegraphics[width=\linewidth]{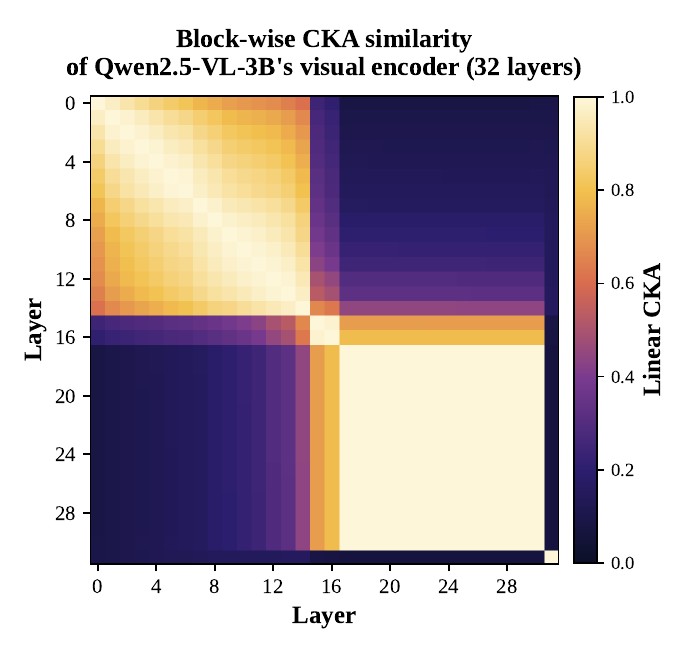}
        \label{fig:block_cka_q25}
    \end{subfigure}
    \hfill
    \begin{subfigure}[t]{0.49\linewidth}
        \centering
        \includegraphics[width=\linewidth]{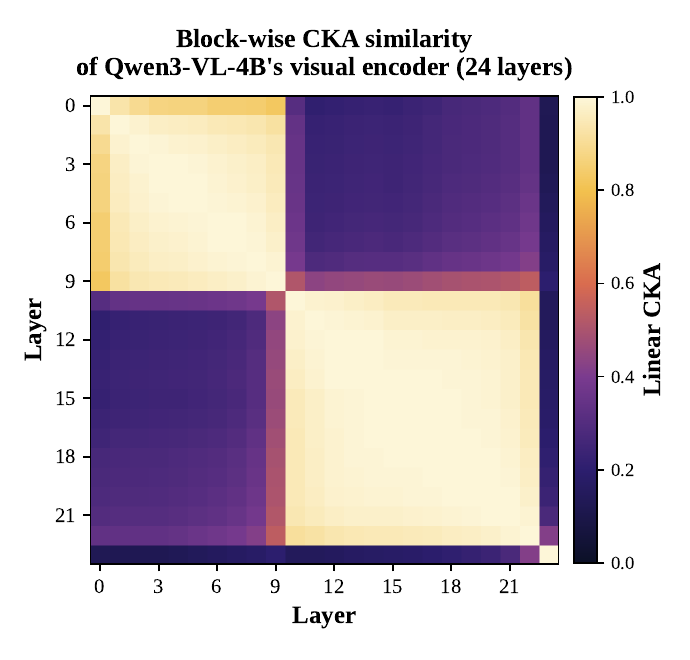}
        \label{fig:block_cka_q3}
    \end{subfigure}
    \vspace{-1.8em}
    \caption{\textbf{Block structure in pretrained visual towers.} Layer-wise linear CKA forms coherent diagonal blocks, indicating contiguous encoder phases with aligned visual transformations. LoopLook places recurrence inside a middle-to-late block, reusing a consistent visual operation across loops.}
    \label{fig:block_cka_main}
\end{figure}

\label{sec:method_prelim}
Given an image $\mathbf{I}$ and an instruction $\mathbf{x}$, a pretrained VLM first encodes the image into visual tokens and then generates the answer conditioned on them:
\begin{equation}
    \mathbf{V}=P_\psi(E_\theta(\mathbf{I})),
    \qquad
    \mathbf{Y}\sim D_\phi(\mathbf{Y}\mid \mathbf{x},\mathbf{V}).
\end{equation}
Here $E_\theta$ is the vision encoder, $P_\psi$ is the projector, and $D_\phi$ is the language model. Once produced, the visual tokens $\mathbf{V}$ remain fixed throughout language generation. Visual perception is therefore effectively a one-shot, task-agnostic decision: the representation must preserve the evidence needed for the downstream task before the model has an opportunity to reason about what the image requires. This can be limiting when useful evidence is fine-grained or degraded.

To address this bottleneck, we allow the visual representation to undergo further encoder-side processing, enabling adaptive and purposeful computation through recurrent perception with shared parameters. LoopLook formulates this process as recursive visual encoding:
\begin{equation}
    \mathbf{H}^{r+1}
    =
    F_{\theta}^{s:e}\!\left(\mathbf{H}^{r};\,\mathbf{x},r\right),
    \quad r=0,\ldots,R_{\max}-1,
    \label{eq:ll_recurrent_update}
\end{equation}
where $R_{\max}$ is the maximum number of loops, $\mathbf{H}^{r}=\{\mathbf{h}^{r}_{i}\}_{i=1}^{N}$ denotes the visual token states at the $r$-th loop, and $F_{\theta}^{s:e}$ is the encoder window reused across loops and conditioned on the instruction $\mathbf{x}$.

\subsection{Recurrent Visual Encoding}
\label{sec:method_recurrent_encoding}

LoopLook inserts recurrence inside the vision encoder. We decompose the encoder into an early stem $E_\theta^{0:s}$, a recurrent window $F_\theta^{s:e}$, and a readout tail $E_\theta^{e:L}$:
\begin{equation}
    E_\theta = E_\theta^{e:L}\circ F_\theta^{s:e}\circ E_\theta^{0:s}.
\end{equation}
The early stem extracts the initial visual state $\mathbf{H}^{0}=E_\theta^{0:s}(\mathbf{I})$. The same window $F_\theta^{s:e}$ is then applied for up to $R_{\max}$ loops, producing a sequence of visual states $\mathbf{H}^{1},\ldots,\mathbf{H}^{R_{\max}}$. The readout tail and projector map each recurrent state back to the standard VLM interface: $\mathbf{V}^{r}=P_\psi\!\left(E_\theta^{e:L}(\mathbf{H}^{r})\right)$.

We place the recurrent window in a middle-to-late region of the encoder, where features move toward semantic integration and become suitable for task-conditioned refinement. Following the block-recurrent view of ViTs~\cite{jacobs2026block}, we treat a coherent CKA phase as a visual \emph{block}: a set of adjacent layers with aligned transformations. Looping within one block gives the recurrent window a consistent operation to reuse. \Cref{fig:block_cka_main} shows this block structure for two Qwen visual towers with complete details in \cref{sec:app_phase_window}.

\subsection{Task-Conditioned Modulation}
\label{sec:method_modulation}
LoopLook further supplies each loop with a compact condition that specifies the task, the current visual state, and the loop step. We first
obtain a task state $\mathbf{q}$ by mean-pooling the instruction input
embeddings from the language model's embedding layer and project the pooled vector to the visual hidden dimension. At loop $r$, we concatenate this task state with the current visual summary and a learnable loop-step embedding $\mathbf{e}_{r}$:
\begin{equation}
    \mathbf{z}^{r}
    = M_{z}\!\left([\mathbf{q};\operatorname{Pool}(\mathbf{H}^{r});\mathbf{e}_{r}]\right).
    \label{eq:ll_modulation_state}
\end{equation}
Here $\mathbf{z}^{r}$ is the recurrent modulation state. The task state gives the visual encoder the instruction context, the pooled visual state summarizes what has been encoded so far, and $\mathbf{e}_{r}$ marks the current loop step.

For each block $\ell$ in the recurrent window, $\mathbf{z}^{r}$ is mapped to affine parameters and applied to normalized visual features:
\begin{equation}
\small
\gamma_{\ell}^{r},\beta_{\ell}^{r}=M_{\ell}(\mathbf{z}^{r}),
    \;
    \operatorname{AdaMod}_{\ell}^{r}(\mathbf{Z})=(1+\gamma_{\ell}^{r})\odot\operatorname{LN}(\mathbf{Z})+\beta_{\ell}^{r}.
    \label{eq:ll_adamod}
\end{equation}
The same recurrent window is therefore reused with a loop-dependent modulation state. 
As the loop unfolds, the visual state evolves, the modulation state is recomputed, and the next update is conditioned on both the instruction and the current visual evidence.

\subsection{Mixture-of-Looks}
\label{sec:method_mixture}

Mixture-of-Looks (MoL) couples adaptive stopping with token-wise state fusion: an exit gate determines the execution depth, and a router combines the available states for each visual token. After each loop, the gate predicts a conditional stopping score:
\begin{equation}
    s^{r}=\sigma\!\left(h_\omega\!\left(\operatorname{Pool}(\mathbf{H}^{r}),\mathbf{q},\mathbf{e}_{r}\right)\right),
    \; r=1,\ldots,R_{\max}.
    \label{eq:ll_exit_score}
\end{equation}
The score $s^{r}$ represents the probability of exiting at loop $r$ after reaching that loop. We set $s^{R_{\max}}=1$ to terminate at the maximum depth. At inference, the encoder exits at the first loop with $s^r>\tau$, yielding an exit depth $\hat R$. The gate scores also define a stopping prior over loops:
\begin{equation}
    \pi_{r}=s^{r}\prod_{t<r}(1-s^{t}),
    \qquad
    \sum_{r=1}^{R_{\max}}\pi_{r}=1.
    \label{eq:ll_stopping_distribution}
\end{equation}
Each prior mass $\pi_r$ is computed from the gate scores observed through loop $r$.

 The token-wise router integrates all $R_{\max}$ recurrent states during training and only the
 executed prefix at inference. The number of available states is $K=R_{\max}$ during training and $K=\hat R$ at inference. The router computes token-specific scores $g_{i,r}=g_\eta(\mathbf{h}_i^r,\mathbf{q},\mathbf{p}_i,r)$, combines them with the stopping prior, and normalizes over these $K$ loops:
\begin{equation}
\begin{gathered}
    \alpha_{i,r}
    =\operatorname{softmax}_{r\le K}\!\left(g_{i,r}+\log(\pi_r+\epsilon)\right),\\
    \tilde{\mathbf{h}}_{i}
    =\sum_{r=1}^{K}\alpha_{i,r}\mathbf{h}_{i}^{r}.
\end{gathered}
\label{eq:ll_mixture_readout}
\end{equation}
Here $\mathbf{p}_{i}$ is the spatial position of token $i$, and $\epsilon>0$ stabilizes the logarithm. The stopping prior provides a sample-level preference over loops, while the token-specific scores allow each visual token to form its own mixture. The fused sequence $\tilde{\mathbf{H}}=\{\tilde{\mathbf{h}}_i\}_{i=1}^{N}$ passes through the readout tail and projector to produce the visual tokens for answer generation:
\begin{equation}
    \mathbf{V}=P_\psi\!\left(E_\theta^{e:L}(\tilde{\mathbf{H}})\right).
    \label{eq:ll_visual_output}
\end{equation}

\subsection{Training Objective}
\label{sec:method_training}

During training, we unroll all $R_{\max}$ loops and retain differentiable fusion across their states, allowing the answer loss to jointly train recurrent refinement, routing, and halting. This separates full-unroll optimization from early-exit inference, following the design principle of differentiable adaptive computation~\cite{eyzaguirre2020differentiable,banino2021pondernet}. We optimize the autoregressive next-token prediction loss together with an expected-depth penalty. Given target response tokens $\mathbf{Y}=\{y_t\}_{t=1}^{T_y}$, where $T_y$ is the response length, the objective is
\begin{equation}
\begin{gathered}
\mathcal{L}
= -\sum_{t=1}^{T_y}
\log p_\phi\!\left(
y_t \mid y_{<t}, \mathbf{x}, \mathbf{V}
\right)
+ \lambda_d \,\mathbb{E}_{\boldsymbol{\pi}}[r],
\\[-2pt]
\mathbb{E}_{\boldsymbol{\pi}}[r]
= \sum_{r=1}^{R_{\max}} r\,\pi_r.
\end{gathered}
\label{eq:ll_training_objective}
\end{equation}
The next-token loss jointly trains the recurrent visual states, the token-wise router, and the exit gate, while the depth term encourages a lower expected loop count under $\boldsymbol{\pi}$.

We gradually introduce the depth penalty using a linear ramp $\lambda_d(u)=\lambda_d^{\star}\cdot\min\!\left(1,\max\!\left(0,\frac{u-t_w}{t_r}\right)\right)$, where $u$ is the training step, $t_w$ is the warmup length, $t_r$ is the ramp length, and $\lambda_d^{\star}$ is the target depth penalty. Both lengths are specified relative to the total number of training steps $T_{\mathrm{train}}$. This schedule first emphasizes learning useful visual updates and then introduces the cost of recurrent depth.
Appendix~\ref{sec:app_algorithm} summarizes the training and inference workflows.

\section{Experiments}
\label{sec:experiments}

\subsection{Experimental Setup}
\label{sec:exp_setup}

\paragraph{Implementation Details.}
We implement experiments on four pretrained VLMs: Qwen2.5-VL-3B~\cite{bai2025qwen25vltechnicalreport}, Qwen3-VL-4B~\cite{bai2025qwen3}, InternVL3.5-4B~\cite{wang2025internvl3}, and LLaVA-1.5-7B~\cite{liu2024improved}, placing the recurrent window in the phase-identified middle-to-late visual blocks (layers 17–30, 10–22, 12–22, and 11–19, respectively). We train LoopLook on 100K Visual CoT~\cite{shao2024visual} samples, using LoRA~\cite{hu2021lora} with a rank of \(r=8\) and a scaling factor of \(\alpha=16\) for the vision encoder and fully training the newly introduced LoopLook modules. With the language model frozen, this adaptation trains only 14.9M, 10.6M, 10.1M, and 11.1M parameters respectively, corresponding to approximately 0.16\%–0.40\% of their total parameters. Each backbone completes training in roughly 1.5--5.5 hours on eight RTX 5090 GPUs. We set the maximum loop depth to $R_{\max}=4$ and use an exit threshold of $0.4$ at inference time. The depth penalty uses a linear ramp with $\lambda_d^{\star}=0.0005$, $t_w=0.6T_{\mathrm{train}}$, and $t_r=0.4T_{\mathrm{train}}$, where $T_{\mathrm{train}}$ denotes the total number of training steps. 
Full details are provided in Appendix~\ref{sec:app_impl}.

\vspace{-1em}
\paragraph{Baselines and Benchmarks.}
We report performance across four backbones and reasoning-side prompting/decoding strategies, including Base direct answering, CoT prompting~\cite{wei2022chain}, and Self-Consistency with five samples~\cite{wang2022self}. We use LLaVA-1.5-7B as the primary setting to compare LoopLook with existing VLM visual-side enhancement methods, including Dense Connector~\cite{yao2024dense} (multi-layer feature concatenation), TGIF~\cite{lin2026text} (instruction-guided layer fusion), and DeepStack~\cite{meng2024deepstack} (anyres multi-crop encoding with intermediate-layer injection).
We evaluate LoopLook on eight extensive downstream benchmarks covering three main abilities: general understanding (MMStar~\cite{chen2024we}, VMCBench~\cite{zhang2025automatedgenerationchallengingmultiplechoice}, AI2D~\cite{10.1007/978-3-319-46493-0_15}), perception and reasoning (V$^\ast$~\cite{wu2024v}, BabyVision~\cite{chen2026babyvision}, VisualPuzzles~\cite{song2025visualpuzzlesdecouplingmultimodalreasoning}), and hallucination robustness (POPE~\cite{li2023evaluating}, HallusionBench~\cite{guan2024hallusionbench}).  Full details are provided in  Appendix ~\ref{sec:app_baselines}.

\subsection{Main Results}
\label{sec:exp_main}

\begin{table*}[t]
    \centering
    \caption{Performance across backbones and traditional prompting/decoding strategies on eight benchmarks.}
    \label{tab:main_results}

    \scriptsize
    \setlength{\tabcolsep}{3pt}
    \renewcommand{\arraystretch}{1.12}

    \begin{tabularx}{1.0\textwidth}{
        @{}
        >{\centering\arraybackslash}p{1.65cm}
        l
        *{9}{>{\centering\arraybackslash}X}
        @{}
    }
        \toprule
        \multirow{2}{*}{\textbf{Backbone}}
        & \multirow{2}{*}{\textbf{Method}}
        & \multicolumn{3}{c}{\textbf{General Understanding}}
        & \multicolumn{3}{c}{\textbf{Perception and Reasoning}}
        & \multicolumn{2}{c}{\textbf{Hallucination Robustness}}
        & \multirow{2}{*}{\textbf{Avg.}} \\
        \cmidrule(lr){3-5}
        \cmidrule(lr){6-8}
        \cmidrule(lr){9-10}
        & &
        \makecell{MMStar} &
        \makecell{VMCBench} &
        \makecell{AI2D} &
        \makecell{V$^\ast$} &
        \makecell{BabyVision} &
        \makecell{VisualPuzzles} &
        \makecell{POPE} &
        \makecell{HalluBench} &
        \\

        \midrule

        % Qwen2.5-VL-3B
        \multirow{4}{*}{\makecell[c]{\textbf{Qwen2.5}\\\textbf{VL-3B}}}
        & Base
        & 55.9 & 74.9 & 81.5
        & \underline{72.8} & 14.9 & 31.2
        & 88.8 & 54.3 & 59.3 \\
        & CoT
        & \underline{56.2}\accgood{0.3} & 73.5\accbad{1.4} & 82.8\accgood{1.3}
        & 65.4\accbad{7.4} & 13.9\accbad{1.0} & 30.1\accbad{1.1}
        & \underline{89.0}\accgood{0.2} & 54.5\accgood{0.2} & 58.2\accbad{1.1} \\
        & SC@5
        & 54.6\accbad{1.3} & \underline{76.3}\accgood{1.4} & \underline{83.9}\accgood{2.4}
        & 67.0\accbad{5.8} & \underline{16.5}\accgood{1.6} & \underline{31.8}\accgood{0.6}
        & 88.6\accbad{0.2} & \underline{58.7}\accgood{4.4} & \underline{59.7}\accgood{0.4} \\
        \rowcolor{LLSoft}
        & \textbf{LoopLook}
        & \textbf{64.5}\accgood{8.6} & \textbf{78.8}\accgood{3.9} & \textbf{84.3}\accgood{2.8}
        & \textbf{78.5}\accgood{5.7} & \textbf{18.6}\accgood{3.7} & \textbf{34.4}\accgood{3.2}
        & \textbf{89.5}\accgood{0.7} & \textbf{62.4}\accgood{8.1} & \textbf{63.9}\accgood{4.6} \\

        \midrule

        % Qwen3-VL-4B
        \multirow{4}{*}{\makecell[c]{\textbf{Qwen3}\\\textbf{VL-4B}}}
        & Base
        & 69.2 & 76.7 & \underline{82.6}
        & 74.9 & \textbf{17.8} & 34.1
        & \underline{90.1} & \underline{60.8} & \underline{63.3} \\
        & CoT
        & \underline{70.6}\accgood{1.4} & 76.9\accgood{0.2} & 82.0\accbad{0.6}
        & 74.3\accbad{0.6} & 13.4\accbad{4.4} & \underline{35.1}\accgood{1.0}
        & 86.9\accbad{3.2} & 54.9\accbad{5.9} & 61.8\accbad{1.5} \\
        & SC@5
        & 70.3\accgood{1.1} & \underline{78.4}\accgood{1.7} & 82.1\accbad{0.5}
        & \underline{76.4}\accgood{1.5} & \underline{15.5}\accbad{2.3} & 33.9\accbad{0.2}
        & 88.6\accbad{1.5} & 60.4\accbad{0.4} & 63.2\accbad{0.1} \\
        \rowcolor{LLSoft}
        & \textbf{LoopLook}
        & \textbf{73.7}\accgood{4.5} & \textbf{81.3}\accgood{4.6} & \textbf{84.8}\accgood{2.2}
        & \textbf{79.6}\accgood{4.7} & 14.4\accbad{3.4} & \textbf{37.0}\accgood{2.9}
        & \textbf{90.8}\accgood{0.7} & \textbf{63.1}\accgood{2.3} & \textbf{65.6}\accgood{2.3} \\

        \midrule

        % InternVL3.5-4B
        \multirow{4}{*}{\makecell[c]{\textbf{InternVL}\\\textbf{3.5-4B}}}
        & Base
        & 65.0 & 78.2 & 82.6
        & 78.0 & 14.9 & \underline{30.3}
        & \underline{88.9} & \underline{59.2} & 62.1 \\
        & CoT
        & \underline{66.9}\accgood{1.9} & \underline{80.8}\accgood{2.6} & \textbf{84.9}\accgood{2.3}
        & \underline{78.5}\accgood{0.5} & 17.5\accgood{2.6} & 27.5\accbad{2.8}
        & 87.4\accbad{1.5} & 58.8\accbad{0.4} & \underline{62.8}\accgood{0.7} \\
        & SC@5
        & 66.7\accgood{1.7} & 79.2\accgood{1.0} & \underline{84.6}\accgood{2.0}
        & 77.0\accbad{1.0} & \underline{17.8}\accgood{2.9} & 27.3\accbad{3.0}
        & 88.5\accbad{0.4} & 57.3\accbad{1.9} & 62.3\accgood{0.2} \\
        \rowcolor{LLSoft}
        & \textbf{LoopLook}
        & \textbf{69.6}\accgood{4.6} & \textbf{81.1}\accgood{2.9} & \underline{84.6}\accgood{2.0}
        & \textbf{81.7}\accgood{3.7} & \textbf{19.6}\accgood{4.7} & \textbf{33.8}\accgood{3.5}
        & \textbf{90.2}\accgood{1.3} & \textbf{61.5}\accgood{2.3} & \textbf{65.3}\accgood{3.2} \\

        \midrule

        % LLaVA-1.5-7B
        \multirow{4}{*}{\makecell[c]{\textbf{LLaVA}\\\textbf{1.5-7B}}}
        & Base
        & \underline{32.3} & 53.8 & 53.7
        & 49.7 & 10.8 & 23.9
        & 89.7 & 47.3 & 45.2 \\
        & CoT
        & 30.4\accbad{1.9} & \underline{55.2}\accgood{1.4} & \underline{54.0}\accgood{0.3}
        & 48.2\accbad{1.5} & 12.9\accgood{2.1} & 24.4\accgood{0.5}
        & \underline{90.1}\accgood{0.4} & 44.6\accbad{2.7} & 45.0\accbad{0.2} \\
        & SC@5
        & 30.8\accbad{1.5} & \textbf{56.5}\accgood{2.7} & 51.7\accbad{2.0}
        & \underline{51.8}\accgood{2.1} & \underline{14.4}\accgood{3.6} & \underline{24.6}\accgood{0.7}
        & 88.3\accbad{1.4} & \underline{48.9}\accgood{1.6} & \underline{45.9}\accgood{0.7} \\
        \rowcolor{LLSoft}
        & \textbf{LoopLook}
        & \textbf{39.2}\accgood{6.9} & 55.1\accgood{1.3} & \textbf{55.8}\accgood{2.1}
        & \textbf{54.5}\accgood{4.8} & \textbf{17.0}\accgood{6.2} & \textbf{24.8}\accgood{0.9}
        & \textbf{90.2}\accgood{0.5} & \textbf{51.4}\accgood{4.1} & \textbf{48.5}\accgood{3.3} \\

        \bottomrule
    \end{tabularx}
\end{table*}

\begin{table*}[t]
    \centering
    \caption{Performance comparison against existing VLM enhancement methods. The first three columns summarize whether each method supports adaptive computation, task guidance, and depth recurrence. Since all existing baselines provide LLaVA-based checkpoints, we use LLaVA-1.5-7B as the base model in this experiment.}
    \label{tab:visual_side_comparison}
    \scriptsize
    \setlength{\tabcolsep}{3pt}
    \renewcommand{\arraystretch}{1.08}
    \begin{tabularx}{\textwidth}{@{}lccc*{8}{>{\centering\arraybackslash}X}c@{}}
        \toprule
        \textbf{Method}
        & \makecell{\textbf{Adaptive}\\\textbf{computation}}
        & \makecell{\textbf{Task-}\\\textbf{guided}}
        & \makecell{\textbf{Depth}\\\textbf{recurrence}}
        & \makecell{MMStar}
        & \makecell{VMC\\Bench}
        & \makecell{AI2D}
        & \makecell{V$^\ast$}
        & \makecell{Baby\\Vision}
        & \makecell{Visual\\Puzzles}
        & \makecell{POPE}
        & \makecell{Hallusion\\Bench}
        & \textbf{Avg.} \\
        \midrule
        Base
        & \xmark & \xmark & \xmark
        & 32.3 & \underline{53.8} & 53.7 & 49.7
        & 10.8 & 23.9 & \underline{89.7} & 47.3 & 45.2 \\
        Dense Connector
        & \xmark & \xmark & \xmark
        & 27.9\accbad{4.4} & 46.6\accbad{7.2} & 50.6\accbad{3.1} & 44.5\accbad{5.2}
        & \underline{13.7}\accgood{2.9} & 23.2\accbad{0.7} & 77.2\accbad{12.5} & \underline{49.3}\accgood{2.0} & 41.6\accbad{3.6} \\
        TGIF
        & \xmark & \cmark & \xmark
        & \underline{33.9}\accgood{1.6} & 52.8\accbad{1.0} & \underline{56.7}\accgood{3.0} & \textbf{55.0}\accgood{5.3}
        & 10.3\accbad{0.5} & \underline{25.0}\accgood{1.1} & 89.6\accbad{0.1} & 46.2\accbad{1.1} & \underline{46.2}\accgood{1.0} \\
        DeepStack
        & \xmark & \xmark & \xmark
        & 31.7\accbad{0.6} & 49.4\accbad{4.4} & \textbf{60.6}\accgood{6.9} & 49.2\accbad{0.5}
        & 9.8\accbad{1.0} & \textbf{25.3}\accgood{1.4} & 86.7\accbad{3.0} & 45.3\accbad{2.0} & 44.8\accbad{0.4} \\
        \rowcolor{LLSoft}
        \textbf{LoopLook}
        & \cmark & \cmark & \cmark
        & \textbf{39.2}\accgood{6.9} & \textbf{55.1}\accgood{1.3} & 55.8\accgood{2.1} & \underline{54.5}\accgood{4.8}
        & \textbf{17.0}\accgood{6.2} & 24.8\accgood{0.9} & \textbf{90.2}\accgood{0.5} & \textbf{51.4}\accgood{4.1} & \textbf{48.5}\accgood{3.3} \\
        \bottomrule
    \end{tabularx}
\end{table*}

\Cref{tab:main_results} shows that LoopLook consistently improves average performance across all four backbones, with gains of up to $4.6$\%. On Qwen2.5-VL-3B, the improvements are particularly pronounced on MMStar and HallusionBench, rising by $8.6$ and $8.1$\%, respectively, with a further $5.7$\% gain on V$^\ast$. CoT and self-consistency often fail to improve fine-grained perception, suggesting that additional language-side computation alone is insufficient when visual evidence requires task-conditioned refinement. We further compare LoopLook with existing VLM enhancement methods using LLaVA-1.5-7B in \Cref{tab:visual_side_comparison}. LoopLook achieves the highest average score of $48.5$, improving the baseline by $3.3$\%, and delivers the strongest performance on five
benchmarks: MMStar, VMCBench, BabyVision, POPE, and HallusionBench. Notably, its gains reach $6.9$\% on MMStar and $6.2$\% on BabyVision, while HallusionBench improves by $4.1$\%. Overall, these results demonstrate the effectiveness of combining question-guidance and adaptive recurrence for refining visual evidence and improving perception across diverse VLM backbones.

\begin{figure*}[t]
    \centering
    \includegraphics[width=\textwidth]{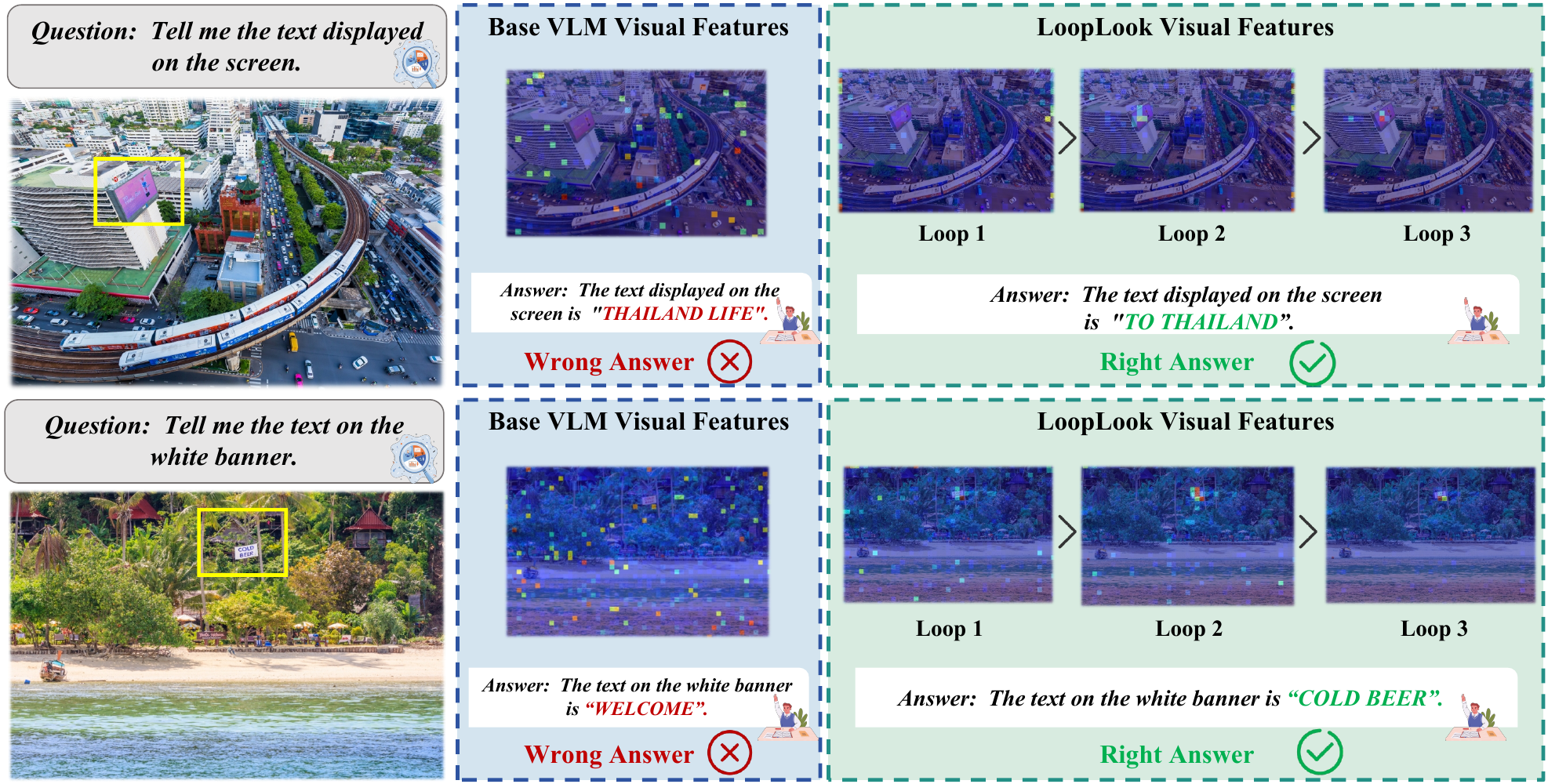}
    \caption{\textbf{Attention heatmap visualization of progressive visual refinement.} 
We visualize the penultimate encoder layer of Qwen2.5-VL-3B at the end of each loop, providing a comparable view of the evolving visual state, and compare the base VLM with LoopLook on V$^\ast$ cases. The base encoder produces diffuse attention heatmaps and incorrect answers, while LoopLook progressively concentrates attention on task-relevant evidence across recurrent loops. }
    \label{fig:vstarcase}
\end{figure*}

\subsection{Analysis}
\label{sec:exp_analysis}

\subsubsection{Adaptive Looping Behaviors}
\label{sec:exp_behaviors}

\paragraph{Progressive Refinement of Visual Attention.}
\Cref{fig:vstarcase} visualizes attention heatmaps of the base VLM and LoopLook on V$^\ast$ examples. Without question guidance, the original visual encoder favors general semantic features, resulting in diffuse attention that can overlook small, task-relevant details. By conditioning recurrent perception on the input question, LoopLook focuses attention on relevant visual regions. Across successive recurrent looks, attention is further deliberately refined and becomes increasingly concentrated on the queried evidence, enabling more precise localization for answer generation.
\begin{table}[t]
    \centering
    \caption{Robustness to reduced resolution on MMStar. We report accuracy for six categories and overall, along with the average number of visual loops. Base denotes a single pass through the visual block.}
    \label{tab:low_resolution_mmstar}
    \scriptsize
    \setlength{\tabcolsep}{1.8pt}
    \renewcommand{\arraystretch}{1.12}
    \resizebox{\columnwidth}{!}{%
    \begin{tabular}{@{}lc*{8}{c}@{}}
        \toprule
        \textbf{Resolution} & \textbf{Method}
        & \textbf{CP} & \textbf{FP} & \textbf{IR} & \textbf{LR}
        & \textbf{Math} & \textbf{S\&T} & \textbf{Overall}
        & \makecell{\textbf{Loops}} \\
        \midrule
        & Base
        & 61.2 & 37.2 & 61.2 & 48.4
        & 50.0 & 33.6 & 48.6 & 1.0 \\
        \rowcolor{LLSoft}
        \cellcolor{white}\multirow{-2}{*}{50K} & \textbf{LoopLook}
        & 68.4 & 49.6 & 63.2 & 55.2
        & 54.0 & 39.2 & 54.9 & 2.3 \\
        \midrule
        & Base
        & 64.4 & 46.0 & 61.2 & 48.4
        & 53.2 & 36.0 & 51.5 & 1.0 \\
        \rowcolor{LLSoft}
        \cellcolor{white}\multirow{-2}{*}{200K} & \textbf{LoopLook}
        & 68.4 & 50.4 & 64.4 & 56.4
        & 53.2 & 38.4 & 55.2 & 1.8 \\
        \midrule
        & Base
        & 68.8 & 45.6 & 64.4 & 54.8
        & 54.4 & 39.6 & 54.6 & 1.0 \\
        \rowcolor{LLSoft}
        \cellcolor{white}\multirow{-2}{*}{450K} & \textbf{LoopLook}
        & 69.6 & 59.6 & 68.0 & 59.2
        & 60.8 & 45.2 & 60.4 & 1.6 \\
        \bottomrule
    \end{tabular}%
    }
    \par\vspace{2pt}
    \parbox{\columnwidth}{\scriptsize
        CP: Coarse Perception; FP: Fine-grained Perception;
        IR: Instance Reasoning; LR: Logical Reasoning;
        S\&T: Science \& Technology.
    }
\end{table}
\paragraph{Low-Resolution Robustness.}
We next stress-test LoopLook by reducing the input resolution, which makes local details harder to preserve in a single encoder pass. \Cref{tab:low_resolution_mmstar} studies whether recurrent visual depth can mitigate the loss of visual evidence under this degradation. On Qwen2.5-VL-3B, LoopLook retains higher MMStar accuracy than the single-pass base at all tested resolutions, with the largest margin at the lowest resolution: at 50K, LoopLook retains $54.9$ overall accuracy compared with $48.6$ for Base, and the margin reaches $12.4$\% on fine-grained perception. The average number of executed loops also decreases as resolution increases, from $2.3$ at 50K to $1.8$ at 200K and $1.6$ at 450K. This behavior echoes human vision, in which insufficiently clear visual evidence can prompt further inspection, whereas sufficiently informative views require less additional scrutiny. Similarly,  LoopLook allocates more computation when visual evidence is weak and exit earlier when the input provides enough detail.

\begin{figure}[t]
    \centering
    \includegraphics[width=\columnwidth]{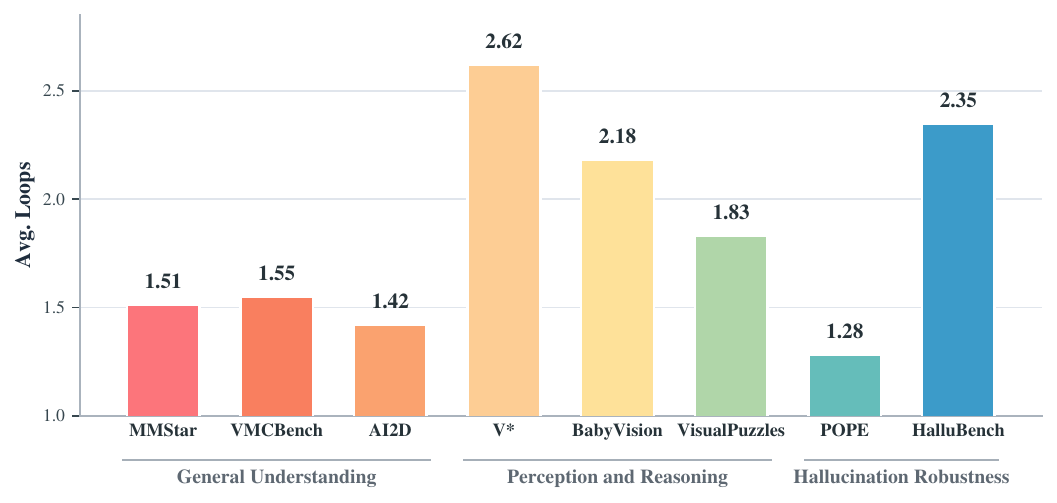}
    \caption{\textbf{Adaptive loop-depth allocation across benchmarks.} LoopLook assigns more recurrent visual processing to fine-grained and hallucination-prone benchmarks, while easier or more direct recognition tasks exit earlier.}
    \label{fig:loop_depth_allocation}
\end{figure}

\paragraph{Loop-Depth Allocation.}
\Cref{fig:loop_depth_allocation} reports the average number of executed loops across benchmarks. The learned exit policy assigns the largest visual depth to V$^\ast$ ($2.62$), HallusionBench ($2.35$), and BabyVision ($2.18$), where answers often depend on small details, visual traps, or weak localized evidence. POPE ($1.28$), AI2D ($1.42$), MMStar ($1.51$), and VMCBench ($1.55$) exit earlier on average, suggesting that many samples in these benchmarks can be resolved with limited additional visual processing. This pattern supports the intended behavior of LoopLook: recurrent computation is used selectively, with harder perception cases receiving more encoder-side depth.

\begin{figure}[t]
    \centering
    \includegraphics[width=1\columnwidth]{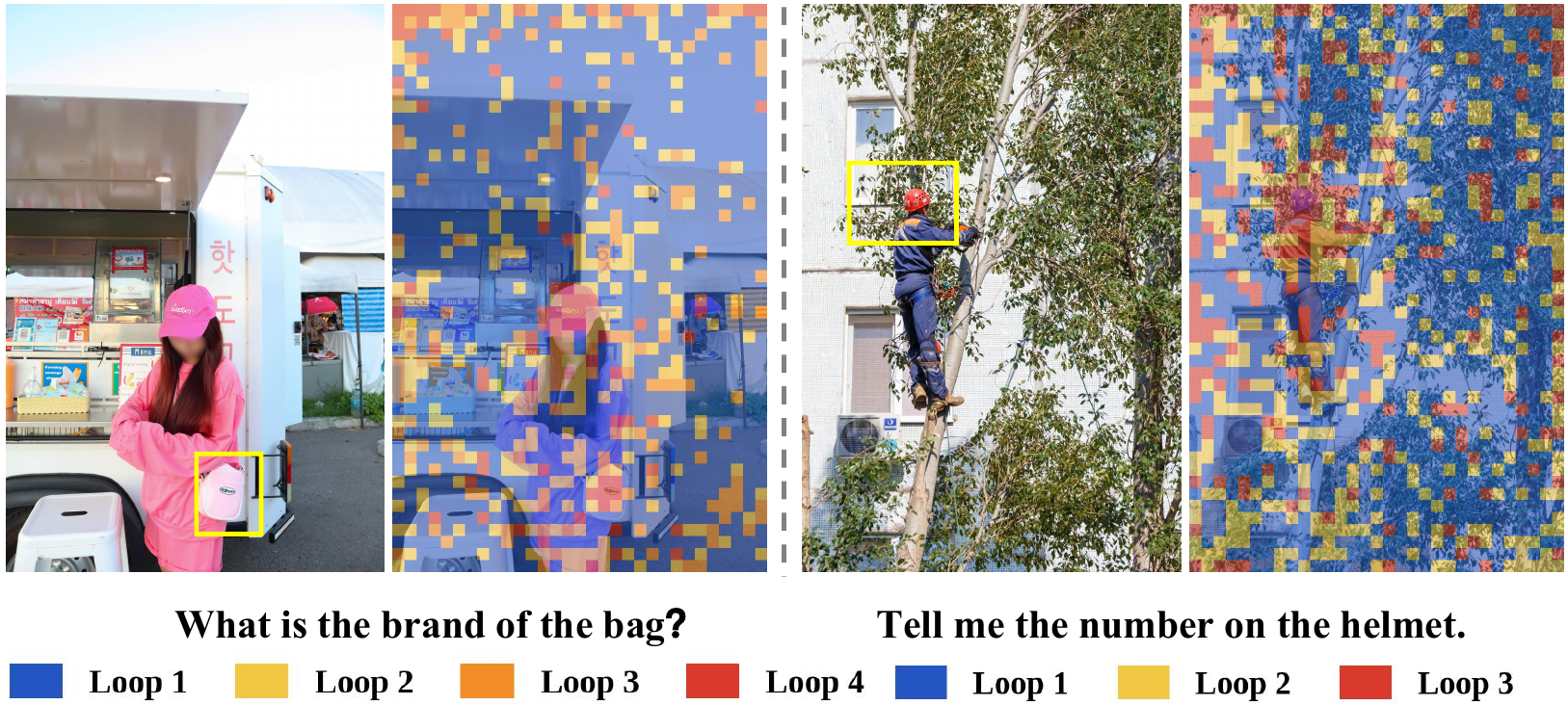}
    \caption{\textbf{Token-wise routing patterns in Mixture-of-Looks.} For each image token, we visualize the loop state with the largest routing weight. Earlier loops cover broad background regions, while later-loop assignments appear around local evidence required by the question, such as the bag region and the helmet area.}
    \label{fig:loopcase}
\end{figure}

\paragraph{Token-wise Routing Patterns.}
\Cref{fig:loopcase} visualizes the token-wise Mixture-of-Looks decision by assigning each image token to the loop state with the largest routing weight. The routing maps show that many background tokens are read from early recurrent states, while tokens near question-relevant regions are more frequently routed to later states. This behavior complements the sample-level exit policy: after deciding how many loops to execute for an image, LoopLook still selects visual depth separately for each token, allowing localized evidence to benefit from deeper recurrent representations without forcing the entire image representation to rely on the last loop.

\subsubsection{Computational Efficiency}
\label{sec:exp_efficiency}

\paragraph{Runtime Overhead.}
\begin{table}[t]
    \centering
    \caption{Runtime overhead evaluated on Qwen2.5-VL-3B. We report time to first token (TTFT), per-token decoding latency, and end-to-end (E2E) runtime for generating 64 tokens.}
    \label{tab:runtime_overhead}
    \scriptsize
    \setlength{\tabcolsep}{3.5pt}
    \renewcommand{\arraystretch}{1.12}
    \resizebox{\columnwidth}{!}{%
    \begin{tabular}{@{}llcccc@{}}
        \toprule
        \textbf{Resolution} & \textbf{Method}
        & \textbf{TTFT}
        & \makecell{\textbf{Overhead}}
        & \textbf{Decode}
        & \textbf{E2E} \\
        \midrule
        & Base
        & 67.2 ms & -- & 21.6 ms/tok & 1.425 s \\
        \rowcolor{LLSoft}
        \cellcolor{white}\multirow{-2}{*}{300K}
        & \textbf{LoopLook}
        & 71.6 ms & $+6.5\%$ & 21.7 ms/tok & 1.435 s \\
        \midrule
        & Base
        & 84.9 ms & -- & 21.8 ms/tok & 1.459 s \\
        \rowcolor{LLSoft}
        \cellcolor{white}\multirow{-2}{*}{550K}
        & \textbf{LoopLook}
        & 89.2 ms & $+5.1\%$ & 21.5 ms/tok & 1.447 s \\
        \bottomrule
    \end{tabular}%
    }
\end{table}

We measure inference latency on a single NVIDIA RTX 5090 using 200 randomly sampled examples across downstream benchmarks. Before timing, we warm up each model with five forward passes and evaluate Base and LoopLook in an interleaved order to mitigate thermal drift or measurement bias.
LoopLook allocates additional computation only to a window of the visual encoder, which contributes a limited fraction of the overall VLM computation while playing a critical role in visual perception, leaving autoregressive decoding unchanged.
As shown in \Cref{tab:runtime_overhead}, this design increases TTFT by only $5.1\%$-$6.5\%$.
The decoding latency remains essentially unchanged, and the end-to-end latency is nearly identical because decoding dominates the total generation time.
This makes encoder-side recurrence an efficient place to spend additional computation: it improves the quality of visual evidence before generation while adding only a small prefill-side cost.

\begin{figure}[t]
    \centering
    \includegraphics[width=\columnwidth]{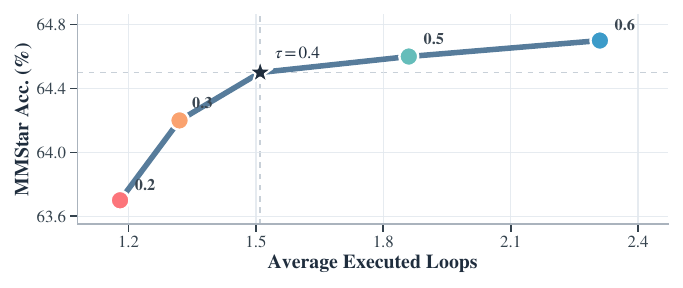}
    \caption{\textbf{Accuracy--compute tradeoff under exit thresholds.} Each point corresponds to one exit threshold on MMStar.}
    \label{fig:accuracy_compute_pareto}
\end{figure}
\paragraph{Accuracy--Compute Tradeoff.}
\Cref{fig:accuracy_compute_pareto} shows the accuracy--compute frontier induced by the exit threshold. A higher threshold makes early exit harder, increasing the average number of executed loops and gradually improving MMStar accuracy. The curve saturates after the default threshold $\tau=0.4$: moving from $\tau=0.4$ to larger thresholds adds more recurrent visual computation but yields only marginal accuracy gains. We therefore use $0.4$ as the default setting, which preserves most of the performance benefit while avoiding unnecessary visual depth.

\begin{table}[t]
    \centering
    \caption{Component ablations and fixed-depth comparisons. Avg. Loops denotes the sample-weighted average number of loops.}
    \label{tab:component_ablations}
    \scriptsize
    \setlength{\tabcolsep}{4pt}
    \renewcommand{\arraystretch}{1.15}
    \resizebox{\columnwidth}{!}{%
    \begin{tabular}{@{}lccccc@{}}
        \toprule
        \textbf{Variant}
        & \textbf{MMStar}
        & \textbf{V$^\ast$}
        & \textbf{HallusionBench}
        & \textbf{Avg.}
        & \textbf{Loops} \\
        \midrule
        \rowcolor{LLSoft}
        \textbf{LoopLook (Full)}
        & \textbf{64.5} & \underline{78.5} & \underline{62.4}
        & \underline{68.5} & 1.89 \\
        Param.-matched
        & 57.2\accbad{7.3} & 73.8\accbad{4.7} & 53.6\accbad{8.8} & 61.5\accbad{7.0} & 1.00 \\
        FLOPs-matched
        & 58.6\accbad{5.9} & 73.8\accbad{4.7} & 58.7\accbad{3.7}
        & 63.7\accbad{4.8} & 1.00 \\
        w/o recurrence
        & 59.7\accbad{4.8} & 74.9\accbad{3.6} & 57.8\accbad{4.6} & 64.1\accbad{4.4} & 1.00 \\
        w/o task condition
        & 58.8\accbad{5.7} & 73.8\accbad{4.7} & 58.7\accbad{3.7} & 63.8\accbad{4.7} & 2.12 \\
        w/o token fusion
        & 62.2\accbad{2.3} & 76.4\accbad{2.1} & 60.0\accbad{2.4} & 66.2\accbad{2.3} & 1.86 \\
        \midrule
        Fixed-1
        & 54.8\accbad{9.7} & 74.9\accbad{3.6} & 59.9\accbad{2.5} & 63.2\accbad{5.3} & 1.00 \\
        Fixed-2
        & 63.2\accbad{1.3} & \underline{78.5} & 62.0\accbad{0.4} & 67.9\accbad{0.6} & 2.00 \\
        Fixed-3
        & 63.8\accbad{0.7} & 77.5\accbad{1.0} & \textbf{62.7}\accgood{0.3} & 68.0\accbad{0.5} & 3.00 \\
        Fixed-4
        & \underline{64.3}\accbad{0.2} & \textbf{81.2}\accgood{2.7} & 61.8\accbad{0.6}
        & \textbf{69.1}\accgood{0.6} & 4.00 \\
        \bottomrule
    \end{tabular}%
    }
\end{table}

\subsection{Ablation Studies}
\label{sec:exp_ablations}

\subsubsection{Contribution of Individual Components}
\Cref{tab:component_ablations} compares fine-tuning baselines, component ablations, and fixed-depth execution  on Qwen2.5-VL-3B. We include parameter- and training-FLOPs-matched fine-tuning, remove recurrence, task conditioning, or token-wise fusion, and compare adaptive execution with fixed budgets of one to four loops. Detailed protocols are given in \cref{sec:app_ablation_protocol}.

Param-Matched FT reaches an average accuracy of $61.5$, while FLOPs-Matched FT reaches $63.7$ after extending the baseline training schedule by $1.5\times$ steps. LoopLook achieves $68.5$, outperforming both fine-tuning baselines.

Removing recurrence reduces the average score by $4.4$ points. Removing task conditioning lowers accuracy by $4.7$ points and increases the average loop count from $1.89$ to $2.12$. Replacing token-wise Mixture-of-Looks with a shared cross-loop readout lowers the average score by $2.3$ points. These results show the contributions of recurrent refinement, question guidance, and token-specific state fusion.

Using the same trained checkpoint, fixed-depth execution reaches average accuracies of $63.2$, $67.9$, and $68.0$ at one, two, and three loops, respectively. Adaptive execution achieves $68.5$ with an average of $1.89$ loops, outperforming both fixed two- and three-loop settings with fewer loops. Fixed four-loop execution reaches $69.1$, but at substantially greater computational depth, highlighting the efficiency of adaptive depth allocation.

\begin{table}[t]
    \centering
    \caption{Recurrent window ablations.}
    \label{tab:window_placement}
    \scriptsize
    \setlength{\tabcolsep}{2.4pt}
    \renewcommand{\arraystretch}{1.03}
    \resizebox{\columnwidth}{!}{%
    \begin{tabular}{@{}lccccc@{}}
        \toprule
        \textbf{Window} & \textbf{Layers}
        & \textbf{MMStar} & \textbf{V$^\ast$}
        & \textbf{HalluBench} & \textbf{Avg.} \\
        \midrule
        \rowcolor{LLSoft}
        \textbf{Mid-Late (Default)} & \textbf{17--30}
        & \textbf{64.5} & \textbf{78.5} & \underline{62.4} & \textbf{68.5} \\
        Cross-block & 14--27
        & 58.4\accbad{6.1} & 74.3\accbad{4.2} & 56.9\accbad{5.5} & 63.2\accbad{5.3} \\
        Early & 0--13
        & 62.0\accbad{2.5} & 77.0\accbad{1.5} & 62.5\accgood{0.1} & 67.2\accbad{1.3} \\
        Early-Mid & 3--16
        & \underline{63.1}\accbad{1.4} & \underline{77.5}\accbad{1.0} & \textbf{62.7}\accgood{0.3} & \underline{67.8}\accbad{0.7} \\
        
        \bottomrule
    \end{tabular}%
    }
\end{table}

\subsubsection{Recurrent Window Placement}

To investigate the effect of recurrent window placement, we sweep the recurrent window while keeping all other settings fixed on Qwen2.5-VL-3B. \Cref{tab:window_placement} shows that the middle-to-late CKA block achieves the best average performance, outperforming Early-Mid, Early, and Cross-block windows by $0.7$, $1.3$, and $5.3$ points, respectively. This suggests that the default window provides a favorable balance: its features are semantically organized enough to respond to the question, yet still retain visual detail that can be refined through recurrent computation. Earlier windows are less effective because they mainly process local appearance and low-level patterns, so recurrence may repeat generic visual extraction and provide weaker task-critical refinement. The cross-block window performs worst, dropping to $63.2$ on average. This supports our phase-aware design: recurrence is most effective when the shared window repeatedly applies a coherent visual transformation. A window that crosses strong CKA boundaries may mix layers with different functional roles, making the repeated update less stable and less aligned with the transformations learned during pretraining.

\section{Conclusion}
We introduced LoopLook, a task-conditioned recurrent visual encoding framework that allocates
additional computation to perception. By combining a shared encoder window with question-guided
modulation, adaptive stopping, and token-wise Mixture-of-Looks, LoopLook refines visual
representations with only lightweight additional training. Experiments across backbones and
benchmarks show consistent improvements in performance, while additional evaluations
demonstrate adaptive depth allocation, robustness to reduced resolution, and modest latency
overhead. Future work could explore finer-grained spatial control and tighter integration with ongoing reasoning to further enhance perception. We hope our findings highlight the value of adaptive visual computation for enhancing multimodal understanding.
\label{sec:conclusion}

{
    \small
    \bibliographystyle{ieeenat_fullname}
    \bibliography{main}

@String(CVPR= {IEEE Conf. Comput. Vis. Pattern Recog.})

@String(ICCV= {Int. Conf. Comput. Vis.})

@String(ECCV= {Eur. Conf. Comput. Vis.})

@String(AAAI = {AAAI})

@String(CVPR  = {CVPR})

@String(ICCV  = {ICCV})

@String(ECCV  = {ECCV})

@inproceedings{jacobs2026block,
  title={Block recurrent dynamics in vision transformers},
  author={Jacobs, Mozes and Fel, Thomas and Hakim, Richard and Brondetta, Alessandra and Ba, Demba and Keller, T Anderson},
  booktitle={International Conference on Learning Representations},
  volume={2026},
  pages={51617--51642},
  year={2026}
}

@inproceedings{liu2024improved,
  title={Improved baselines with visual instruction tuning},
  author={Liu, Haotian and Li, Chunyuan and Li, Yuheng and Lee, Yong Jae},
  booktitle={2024 IEEE/CVF Conference on Computer Vision and Pattern Recognition (CVPR)},
  pages={26286--26296},
  year={2024},
  organization={IEEE}
}

@article{dai2023instructblip,
  title={Instructblip: Towards general-purpose vision-language models with instruction tuning},
  author={Dai, Wenliang and Li, Junnan and Li, Dongxu and Tiong, Anthony and Zhao, Junqi and Wang, Weisheng and Li, Boyang and Fung, Pascale N and Hoi, Steven},
  journal={Advances in neural information processing systems},
  volume={36},
  pages={49250--49267},
  year={2023}
}

@article{meng2024deepstack,
  title={Deepstack: Deeply stacking visual tokens is surprisingly simple and effective for lmms},
  author={Meng, Lingchen and Yang, Jianwei and Tian, Rui and Dai, Xiyang and Wu, Zuxuan and Gao, Jianfeng and Jiang, Yu-Gang},
  journal={Advances in Neural Information Processing Systems},
  volume={37},
  pages={23464--23487},
  year={2024}
}

@article{lin2026text,
  title={Text-Guided Layer Fusion Mitigates Hallucination in Multimodal LLMs},
  author={Lin, Chenchen and Su, Sanbao and Luo, Rachel and Chen, Yuxiao and Wang, Yan and Pavone, Marco and Miao, Fei},
  journal={arXiv preprint arXiv:2601.03100},
  year={2026}
}

@article{yao2024dense,
  title={Dense connector for mllms},
  author={Yao, Huanjin and Wu, Wenhao and Yang, Taojiannan and Song, YuXin and Zhang, Mengxi and Feng, Haocheng and Sun, Yifan and Li, Zhiheng and Ouyang, Wanli and Wang, Jingdong},
  journal={Advances in Neural Information Processing Systems},
  volume={37},
  pages={33108--33140},
  year={2024}
}

@article{wang2022self,
  title={Self-consistency improves chain of thought reasoning in language models},
  author={Wang, Xuezhi and Wei, Jason and Schuurmans, Dale and Le, Quoc and Chi, Ed and Narang, Sharan and Chowdhery, Aakanksha and Zhou, Denny},
  journal={arXiv preprint arXiv:2203.11171},
  year={2022}
}

@article{wei2022chain,
  title={Chain-of-thought prompting elicits reasoning in large language models},
  author={Wei, Jason and Wang, Xuezhi and Schuurmans, Dale and Bosma, Maarten and Xia, Fei and Chi, Ed and Le, Quoc V and Zhou, Denny and others},
  journal={Advances in neural information processing systems},
  volume={35},
  pages={24824--24837},
  year={2022}
}

@inproceedings{guan2024hallusionbench,
  title={Hallusionbench: an advanced diagnostic suite for entangled language hallucination and visual illusion in large vision-language models},
  author={Guan, Tianrui and Liu, Fuxiao and Wu, Xiyang and Xian, Ruiqi and Li, Zongxia and Liu, Xiaoyu and Wang, Xijun and Chen, Lichang and Huang, Furong and Yacoob, Yaser and others},
  booktitle={2024 IEEE/CVF Conference on Computer Vision and Pattern Recognition (CVPR)},
  pages={14375--14385},
  year={2024},
  organization={IEEE}
}

@inproceedings{li2023evaluating,
  title={Evaluating object hallucination in large vision-language models},
  author={Li, Yifan and Du, Yifan and Zhou, Kun and Wang, Jinpeng and Zhao, Xin and Wen, Ji-Rong},
  booktitle={Proceedings of the 2023 conference on empirical methods in natural language processing},
  pages={292--305},
  year={2023}
}

@misc{song2025visualpuzzlesdecouplingmultimodalreasoning,
      title={VisualPuzzles: Decoupling Multimodal Reasoning Evaluation from Domain Knowledge}, 
      author={Yueqi Song and Tianyue Ou and Yibo Kong and Zecheng Li and Graham Neubig and Xiang Yue},
      year={2025},
      eprint={2504.10342},
      archivePrefix={arXiv},
      primaryClass={cs.CL},
      url={https://arxiv.org/abs/2504.10342}, 
}

@article{chen2026babyvision,
  title={Babyvision: Visual reasoning beyond language},
  author={Chen, Liang and Xie, Weichu and Liang, Yiyan and He, Hongfeng and Zhao, Hans and Yang, Zhibo and Huang, Zhiqi and Wu, Haoning and Lu, Haoyu and Bao, Yiping and others},
  journal={arXiv preprint arXiv:2601.06521},
  year={2026}
}

@inproceedings{wu2024v,
  title={V*: Guided visual search as a core mechanism in multimodal llms},
  author={Wu, Penghao and Xie, Saining},
  booktitle={2024 IEEE/CVF Conference on Computer Vision and Pattern Recognition (CVPR)},
  pages={13084--13094},
  year={2024},
  organization={IEEE}
}

@misc{zhang2025automatedgenerationchallengingmultiplechoice,
      title={Automated Generation of Challenging Multiple-Choice Questions for Vision Language Model Evaluation}, 
      author={Yuhui Zhang and Yuchang Su and Yiming Liu and Xiaohan Wang and James Burgess and Elaine Sui and Chenyu Wang and Josiah Aklilu and Alejandro Lozano and Anjiang Wei and Ludwig Schmidt and Serena Yeung-Levy},
      year={2025},
      eprint={2501.03225},
      archivePrefix={arXiv},
      primaryClass={cs.CV},
      url={https://arxiv.org/abs/2501.03225}, 
}

@article{chen2024we,
  title={Are we on the right way for evaluating large vision-language models?},
  author={Chen, Lin and Li, Jinsong and Dong, Xiaoyi and Zhang, Pan and Zang, Yuhang and Chen, Zehui and Duan, Haodong and Wang, Jiaqi and Qiao, Yu and Lin, Dahua and others},
  journal={Advances in Neural Information Processing Systems},
  volume={37},
  pages={27056--27087},
  year={2024}
}

@article{hu2021lora,
  title={Lora: Low-rank adaptation of large language models},
  author={Hu, Edward J and Shen, Yelong and Wallis, Phillip and Allen-Zhu, Zeyuan and Li, Yuanzhi and Wang, Shean and Wang, Lu and Chen, Weizhu},
  journal={arXiv preprint arXiv:2106.09685},
  year={2021}
}

@article{shao2024visual,
  title={Visual cot: Advancing multi-modal language models with a comprehensive dataset and benchmark for chain-of-thought reasoning},
  author={Shao, Hao and Qian, Shengju and Xiao, Han and Song, Guanglu and Zong, Zhuofan and Wang, Letian and Liu, Yu and Li, Hongsheng},
  journal={Advances in Neural Information Processing Systems},
  volume={37},
  pages={8612--8642},
  year={2024}
}

@article{wang2025internvl3,
  title={Internvl3. 5: Advancing open-source multimodal models in versatility, reasoning, and efficiency},
  author={Wang, Weiyun and Gao, Zhangwei and Gu, Lixin and Pu, Hengjun and Cui, Long and Wei, Xingguang and Liu, Zhaoyang and Jing, Linglin and Ye, Shenglong and Shao, Jie and others},
  journal={arXiv preprint arXiv:2508.18265},
  year={2025}
}

@inproceedings{radford2021clip,
  title={Learning transferable visual models from natural language supervision},
  author={Radford, Alec and Kim, Jong Wook and Hallacy, Chris and Ramesh, Aditya and Goh, Gabriel and Agarwal, Sandhini and Sastry, Girish and Askell, Amanda and Mishkin, Pamela and Clark, Jack and others},
  booktitle={International conference on machine learning},
  pages={8748--8763},
  year={2021},
  organization={PmLR}
}

@misc{bai2025qwen25vltechnicalreport,
      title={Qwen2.5-VL Technical Report}, 
      author={Shuai Bai and Keqin Chen and Xuejing Liu and Jialin Wang and Wenbin Ge and Sibo Song and Kai Dang and Peng Wang and Shijie Wang and Jun Tang and Humen Zhong and Yuanzhi Zhu and Mingkun Yang and Zhaohai Li and Jianqiang Wan and Pengfei Wang and Wei Ding and Zheren Fu and Yiheng Xu and Jiabo Ye and Xi Zhang and Tianbao Xie and Zesen Cheng and Hang Zhang and Zhibo Yang and Haiyang Xu and Junyang Lin},
      year={2025},
      eprint={2502.13923},
      archivePrefix={arXiv},
      primaryClass={cs.CV},
      url={https://arxiv.org/abs/2502.13923}, 
}

@inproceedings{li2026latent,
  title={Latent visual reasoning},
  author={Li, Bangzheng and Sun, Ximeng and Liu, Jiang and Wang, Ze and Wu, Jialian and Yu, Xiaodong and Barsoum, Emad and Chen, Muhao and Liu, Zicheng},
  booktitle={International Conference on Learning Representations},
  volume={2026},
  pages={148076--148090},
  year={2026}
}

@article{jeon2026vision,
  title={Vision-aligned latent reasoning for multi-modal large language model},
  author={Jeon, Byungwoo and Jeong, Yoonwoo and Lee, Hyunseok and Cho, Minsu and Shin, Jinwoo},
  journal={arXiv preprint arXiv:2602.04476},
  year={2026}
}

@article{pham2025multimodal,
  title={Multimodal chain of continuous thought for latent-space reasoning in vision-language models},
  author={Pham, Tan-Hanh and Ngo, Chris},
  journal={arXiv preprint arXiv:2508.12587},
  year={2025}
}

@inproceedings{he2025self,
  title={Self-correction is more than refinement: A learning framework for visual and language reasoning tasks},
  author={He, Jiayi and Lin, Hehai and Wang, Qingyun and Fung, Yi R and Ji, Heng},
  booktitle={Findings of the Association for Computational Linguistics: ACL 2025},
  pages={6405--6421},
  year={2025}
}

@inproceedings{cheng2025vision,
  title={Vision-language models can self-improve reasoning via reflection},
  author={Cheng, Kanzhi and YanTao, Li and Xu, Fangzhi and Zhang, Jianbing and Zhou, Hao and Liu, Yang},
  booktitle={Proceedings of the 2025 Conference of the Nations of the Americas Chapter of the Association for Computational Linguistics: Human Language Technologies (Volume 1: Long Papers)},
  pages={8876--8892},
  year={2025}
}

@inproceedings{eyzaguirre2020differentiable,
  title={Differentiable adaptive computation time for visual reasoning},
  author={Eyzaguirre, Cristobal and Soto, Alvaro},
  booktitle={2020 IEEE/CVF Conference on Computer Vision and Pattern Recognition (CVPR)},
  pages={12814--12822},
  year={2020},
  organization={IEEE}
}

@article{wang2026vl,
  title={Vl-rethinker: Incentivizing self-reflection of vision-language models with reinforcement learning},
  author={Wang, Haozhe and Qu, Chao and Huang, Zuming and Chu, Wei and Lin, Fangzhen and Chen, Wenhu},
  journal={Advances in Neural Information Processing Systems},
  volume={38},
  pages={30865--30891},
  year={2026}
}

@InProceedings{pmlr-v306-hu26ba,
  title = 	 {{TVI}-{C}o{T}: Text-Visual Interleaved Chain-of-Thought Reasoning for Multimodal Understanding},
  author =       {Hu, Lianyu and Xiaoyu, Ma and Liao, Zeqin and Liu, Yang},
  booktitle = 	 {Proceedings of the 43rd International Conference on Machine Learning},
  pages = 	 {45820--45831},
  year = 	 {2026},
  editor = 	 {Zhang, Tong and Dudik, Miroslav and Jaggi, Martin and Agarwal, Alekh and Li, Sharon and Schuurmans, Dale and Zhu, Jerry and Berkenkamp, Felix and Dong, Hanze and Bietti, Alberto},
  volume = 	 {306},
  series = 	 {Proceedings of Machine Learning Research},
  month = 	 {06--11 Jul},
  publisher =    {PMLR},
  url = 	 {https://proceedings.mlr.press/v306/hu26ba.html}
}

@InProceedings{10.1007/978-3-319-46493-0_15,
author="Kembhavi, Aniruddha
and Salvato, Mike
and Kolve, Eric
and Seo, Minjoon
and Hajishirzi, Hannaneh
and Farhadi, Ali",
editor="Leibe, Bastian
and Matas, Jiri
and Sebe, Nicu
and Welling, Max",
title="A Diagram is Worth a Dozen Images",
booktitle="Computer Vision -- ECCV 2016",
year="2016",
publisher="Springer International Publishing",
address="Cham",
pages="235--251",
isbn="978-3-319-46493-0"
}

@article{banino2021pondernet,
  title={Pondernet: Learning to ponder},
  author={Banino, Andrea and Balaguer, Jan and Blundell, Charles},
  journal={arXiv preprint arXiv:2107.05407},
  year={2021}
}

@article{LAMME2000571,
title = {The distinct modes of vision offered by feedforward and recurrent processing},
journal = {Trends in Neurosciences},
volume = {23},
number = {11},
pages = {571-579},
year = {2000},
issn = {0166-2236},
doi = {https://doi.org/10.1016/S0166-2236(00)01657-X},
url = {https://www.sciencedirect.com/science/article/pii/S016622360001657X},
author = {Victor A.F. Lamme and Pieter R. Roelfsema}
}

@article{PARR201792,
title = {The active construction of the visual world},
journal = {Neuropsychologia},
volume = {104},
pages = {92-101},
year = {2017},
issn = {0028-3932},
doi = {https://doi.org/10.1016/j.neuropsychologia.2017.08.003},
url = {https://www.sciencedirect.com/science/article/pii/S0028393217302907},
author = {Thomas Parr and Karl J. Friston}
}

@misc{you2026improvingtesttimescalingadaptive,
      title={Improving Test-Time Scaling with Adaptive Looped Transformers}, 
      author={Yichen You and Tianyu Fu and Aosong Feng and Xingtai Lv and Xuefei Ning and Ning Ding and Yu Wang},
      year={2026},
      eprint={2609.35748},
      archivePrefix={arXiv},
      primaryClass={cs.CL},
      url={https://arxiv.org/abs/2609.35748}, 
}

@inproceedings{lin2014microsoft,
  title={Microsoft coco: Common objects in context},
  author={Lin, Tsung-Yi and Maire, Michael and Belongie, Serge and Hays, James and Perona, Pietro and Ramanan, Deva and Doll{\'a}r, Piotr and Zitnick, C Lawrence},
  booktitle={European conference on computer vision},
  pages={740--755},
  year={2014},
  organization={Springer}
}

@inproceedings{caron2021emerging,
  title={Emerging properties in self-supervised vision transformers},
  author={Caron, Mathilde and Touvron, Hugo and Misra, Ishan and J{\'e}gou, Herv{\'e} and Mairal, Julien and Bojanowski, Piotr and Joulin, Armand},
  booktitle={2021 IEEE/CVF international conference on computer vision (ICCV)},
  pages={9630--9640},
  year={2021},
  organization={IEEE}
}

@misc{touvron2023llama2openfoundation,
      title={Llama 2: Open Foundation and Fine-Tuned Chat Models}, 
      author={Hugo Touvron and Louis Martin and Kevin Stone and Peter Albert and Amjad Almahairi and Yasmine Babaei and Nikolay Bashlykov and Soumya Batra and Prajjwal Bhargava and Shruti Bhosale and Dan Bikel and Lukas Blecher and Cristian Canton Ferrer and Moya Chen and Guillem Cucurull and David Esiobu and Jude Fernandes and Jeremy Fu and Wenyin Fu and Brian Fuller and Cynthia Gao and Vedanuj Goswami and Naman Goyal and Anthony Hartshorn and Saghar Hosseini and Rui Hou and Hakan Inan and Marcin Kardas and Viktor Kerkez and Madian Khabsa and Isabel Kloumann and Artem Korenev and Punit Singh Koura and Marie-Anne Lachaux and Thibaut Lavril and Jenya Lee and Diana Liskovich and Yinghai Lu and Yuning Mao and Xavier Martinet and Todor Mihaylov and Pushkar Mishra and Igor Molybog and Yixin Nie and Andrew Poulton and Jeremy Reizenstein and Rashi Rungta and Kalyan Saladi and Alan Schelten and Ruan Silva and Eric Michael Smith and Ranjan Subramanian and Xiaoqing Ellen Tan and Binh Tang and Ross Taylor and Adina Williams and Jian Xiang Kuan and Puxin Xu and Zheng Yan and Iliyan Zarov and Yuchen Zhang and Angela Fan and Melanie Kambadur and Sharan Narang and Aurelien Rodriguez and Robert Stojnic and Sergey Edunov and Thomas Scialom},
      year={2023},
      eprint={2307.09288},
      archivePrefix={arXiv},
      primaryClass={cs.CL},
      url={https://arxiv.org/abs/2307.09288}, 
}

@article{ye2026ebb,
  title={The Ebb and Flow of Multimodal Focus: Scheduling Visual Relay Windows for Grounded VLM Reasoning},
  author={Ye, Wencheng and Bin, Yi and Ding, Yujuan and Fang, Hongye and Wang, Zheng and Xu, Xing and Song, Jingkuan and Zhang, Yun and Da, Sirui and Shen, Heng Tao},
  journal={arXiv preprint arXiv:2607.11436},
  year={2026}
}

@article{wang2026smelt,
  title={Smelt: Scaling laws for compute-matched moe looped transformers},
  author={Wang, Shaowen and Zhang, Ge and Luo, Kairong and Wu, Yuhao and Liu, Shaofan and Liu, Jiaheng and Huang, Wenhao and Yan, Shen and Li, Jian},
  journal={arXiv preprint arXiv:2609.01343},
  year={2026}
}

@article{li2025tokenpacker,
  title={Tokenpacker: Efficient visual projector for multimodal llm},
  author={Li, Wentong and Yuan, Yuqian and Liu, Jian and Tang, Dongqi and Wang, Song and Qin, Jie and Zhu, Jianke and Zhang, Lei},
  journal={International Journal of Computer Vision},
  volume={133},
  number={10},
  pages={6794--6812},
  year={2025},
  publisher={Springer}
}

@article{li2025mini,
  title={Mini-gemini: Mining the potential of multi-modality vision language models},
  author={Li, Yanwei and Zhang, Yuechen and Wang, Chengyao and Zhong, Zhisheng and Chen, Yixin and Chu, Ruihang and Liu, Shaoteng and Jia, Jiaya},
  journal={IEEE Transactions on Pattern Analysis and Machine Intelligence},
  year={2025},
  publisher={IEEE}
}

@inproceedings{li2024monkey,
  title={Monkey: Image resolution and text label are important things for large multi-modal models},
  author={Li, Zhang and Yang, Biao and Liu, Qiang and Ma, Zhiyin and Zhang, Shuo and Yang, Jingxu and Sun, Yabo and Liu, Yuliang and Bai, Xiang},
  booktitle={2024 IEEE/CVF Conference on Computer Vision and Pattern Recognition (CVPR)},
  pages={26753--26763},
  year={2024},
  organization={IEEE}
}

@misc{liu2024llavanext,
    title={LLaVA-NeXT: Improved reasoning, OCR, and world knowledge},
    url={https://llava-vl.github.io/blog/2024-01-30-llava-next/},
    author={Liu, Haotian and Li, Chunyuan and Li, Yuheng and Li, Bo and Zhang, Yuanhan and Shen, Sheng and Lee, Yong Jae},
    month={January},
    year={2024}
}

@InProceedings{pmlr-v306-wu26f,
  title = 	 {From Seeing to Thinking: Decoupling Perception and Reasoning Improves Post-Training of Vision-Language Models},
  author =       {Wu, Juncheng and Chen, Hardy and Tu, Haoqin and Tang, Xianfeng and Shi, Freda and Liu, Hui and Lu, Hanqing and Xie, Cihang and Zhou, Yuyin},
  booktitle = 	 {Proceedings of the 43rd International Conference on Machine Learning},
  pages = 	 {135672--135690},
  year = 	 {2026},
  editor = 	 {Zhang, Tong and Dudik, Miroslav and Jaggi, Martin and Agarwal, Alekh and Li, Sharon and Schuurmans, Dale and Zhu, Jerry and Berkenkamp, Felix and Dong, Hanze and Bietti, Alberto},
  volume = 	 {306},
  series = 	 {Proceedings of Machine Learning Research},
  month = 	 {06--11 Jul},
  publisher =    {PMLR},
  url = 	 {https://proceedings.mlr.press/v306/wu26f.html}
}

@InProceedings{pmlr-v235-lin24c,
  title = 	 {Revisiting the Role of Language Priors in Vision-Language Models},
  author =       {Lin, Zhiqiu and Chen, Xinyue and Pathak, Deepak and Zhang, Pengchuan and Ramanan, Deva},
  booktitle = 	 {Proceedings of the 41st International Conference on Machine Learning},
  pages = 	 {29914--29934},
  year = 	 {2024},
  editor = 	 {Salakhutdinov, Ruslan and Kolter, Zico and Heller, Katherine and Weller, Adrian and Oliver, Nuria and Scarlett, Jonathan and Berkenkamp, Felix},
  volume = 	 {235},
  series = 	 {Proceedings of Machine Learning Research},
  month = 	 {21--27 Jul},
  publisher =    {PMLR},
  url = 	 {https://proceedings.mlr.press/v235/lin24c.html}
}

@article{prisadnikov2025vision,
  title={Vision encoders should be image size agnostic and task driven},
  author={Prisadnikov, Nedyalko and Paudel, Danda Pani and Fu, Yuqian and Van Gool, Luc},
  journal={arXiv preprint arXiv:2508.16317},
  year={2025}
}

@article{hayhoe2005eye,
  title={Eye movements in natural behavior},
  author={Hayhoe, Mary and Ballard, Dana},
  journal={Trends in cognitive sciences},
  volume={9},
  number={4},
  pages={188--194},
  year={2005},
  publisher={Elsevier}
}

@book{yarbus2013eye,
  title={Eye movements and vision},
  author={Yarbus, Alfred L},
  year={2013},
  publisher={Springer}
}

@inproceedings{kim2026tabflash,
  title={Tabflash: Efficient table understanding with progressive question conditioning and token focusing},
  author={Kim, Jongha and Bae, Minseong and Lee, Sanghyeok and Yoon, Jinsung and Kim, Hyunwoo J},
  booktitle={Proceedings of the AAAI Conference on Artificial Intelligence},
  volume={40},
  number={27},
  pages={22573--22581},
  year={2026}
}

@inproceedings{vasu2025fastvlm,
  title={Fastvlm: Efficient vision encoding for vision language models},
  author={Vasu, Pavan Kumar Anasosalu and Faghri, Fartash and Li, Chun-Liang and Koc, Cem and True, Nate and Antony, Albert and Santhanam, Gokula and Gabriel, James and Grasch, Peter and Tuzel, Oncel and others},
  booktitle={Proceedings of the IEEE/CVF Conference on Computer Vision and Pattern Recognition},
  pages={19769--19780},
  year={2025}
}

@inproceedings{liang2015recurrent,
  title={Recurrent convolutional neural network for object recognition},
  author={Liang, Ming and Hu, Xiaolin},
  booktitle={Proceedings of the IEEE conference on computer vision and pattern recognition},
  pages={3367--3375},
  year={2015}
}

@misc{burzio2026d,
      title={D\'ej\`a View: Looping Transformers for Multi-View 3D Reconstruction}, 
      author={Alessandro Burzio and Tobias Fischer and Sven Elflein and Qunjie Zhou and Riccardo de Lutio and Jiawei Ren and Jiahui Huang and Shengyu Huang and Marc Pollefeys and Laura Leal-Taixé and Zan Gojcic and Haithem Turki},
      year={2026},
      eprint={2605.30215},
      archivePrefix={arXiv},
      primaryClass={cs.CV},
      url={https://arxiv.org/abs/2605.30215}, 
}

@article{linsley2018learning,
  title={Learning long-range spatial dependencies with horizontal gated recurrent units},
  author={Linsley, Drew and Kim, Junkyung and Veerabadran, Vijay and Windolf, Charles and Serre, Thomas},
  journal={Advances in neural information processing systems},
  volume={31},
  year={2018}
}

@article{bulat2026one,
  title={One Block, Multiple Depths: Recurrent Vision Transformers with Depth-Programmed Experts},
  author={Bulat, Adrian and Ouali, Yassine and Tzimiropoulos, Georgios},
  journal={arXiv preprint arXiv:2610.12448},
  year={2026}
}

@article{lee2026soft,
  title={Soft Mixture-of-Recursions: Going Deeper with Recursive Vision Transformers},
  author={Lee, Sang In and Park, Jihun},
  journal={arXiv preprint arXiv:2607.00774},
  year={2026}
}

@article{shu2026loopvit,
  title={Loopvit: Scaling visual arc with looped transformers},
  author={Shu, Wen-Jie and Qiu, Xuerui and Zhu, Rui-Jie and Chen, Harold Haodong and Liu, Yexin and Yang, Harry},
  journal={arXiv preprint arXiv:2602.02156},
  year={2026}
}

@article{bae2026mixture,
  title={Mixture-of-recursions: Learning dynamic recursive depths for adaptive token-level computation},
  author={Bae, Sangmin and Kim, Yujin and Bayat, Reza and Kim, Sungnyun and Ha, Jiyoun and Schuster, Tal and Fisch, Adam and Harutyunyan, Hrayr and Ji, Ziwei and Courville, Aaron and others},
  journal={Advances in Neural Information Processing Systems},
  volume={38},
  pages={96572--96617},
  year={2026}
}

@article{geiping2026scaling,
  title={Scaling up test-time compute with latent reasoning: A recurrent depth approach},
  author={Geiping, Jonas and McLeish, Sean and Jain, Neel and Kirchenbauer, John and Singh, Siddharth and Bartoldson, Brian and Kailkhura, Bhavya and Bhatele, Abhinav and Goldstein, Tom},
  journal={Advances in Neural Information Processing Systems},
  volume={38},
  pages={41340--41391},
  year={2026}
}

@article{mcleish2025teaching,
  title={Teaching pretrained language models to think deeper with retrofitted recurrence},
  author={McLeish, Sean and Li, Ang and Kirchenbauer, John and Kalra, Dayal Singh and Bartoldson, Brian R and Kailkhura, Bhavya and Schwarzschild, Avi and Geiping, Jonas and Goldstein, Tom and Goldblum, Micah},
  journal={arXiv preprint arXiv:2511.07384},
  year={2025}
}

@inproceedings{bae2025relaxed,
  title={Relaxed recursive transformers: Effective parameter sharing with layer-wise lora},
  author={Bae, Sangmin and Fisch, Adam and Harutyunyan, Hrayr and Ji, Ziwei and Kim, Seungyeon and Schuster, Tal},
  booktitle={International Conference on Learning Representations},
  volume={2025},
  pages={34282--34327},
  year={2025}
}

@article{zhu2025scaling,
  title={Scaling latent reasoning via looped language models},
  author={Zhu, Rui-Jie and Wang, Zixuan and Hua, Kai and Zhang, Tianyu and Li, Ziniu and Que, Haoran and Wei, Boyi and Wen, Zixin and Yin, Fan and Xing, He and others},
  journal={arXiv preprint arXiv:2510.25741},
  year={2025}
}

@article{graves2016adaptive,
  title={Adaptive computation time for recurrent neural networks},
  author={Graves, Alex},
  journal={arXiv preprint arXiv:1603.08983},
  year={2016}
}

@article{jeddi2026loopformer,
  title={Loopformer: Elastic-depth looped transformers for latent reasoning via shortcut modulation},
  author={Jeddi, Ahmadreza and Ciccone, Marco and Taati, Babak},
  journal={arXiv preprint arXiv:2602.11451},
  year={2026}
}

@article{qian2026loopvl,
  title={LoopVL: Recurrent Visual Intelligence},
  author={Qian, Zhe and Gong, Ziyang and Xu, Zhongxing and Li, Hehan and Wang, Zhonghua and Luo, Fei and Wang, Mingxuan and Yang, Xue and Ma, Yanbiao and Yan, Junchi and others},
  journal={arXiv preprint arXiv:2609.38426},
  year={2026}
}

@article{dehghani2018universal,
  title={Universal transformers},
  author={Dehghani, Mostafa and Gouws, Stephan and Vinyals, Oriol and Uszkoreit, Jakob and Kaiser, {\L}ukasz},
  journal={arXiv preprint arXiv:1807.03819},
  year={2018}
}

@misc{kim2026mol,
      title={Mixture of Layers: Dynamic Layer Routing for Visual Reasoning}, 
      author={Jeonghwan Kim and Sofia Stoica and Jiwan Chung and Ansel Blume and Hyeonjeong Ha and Zhenhailong Wang and Xin Luna Dong and Heng Ji},
      year={2026},
      eprint={2610.09440},
      archivePrefix={arXiv},
      primaryClass={cs.CV},
      url={https://arxiv.org/abs/2610.09440}, 
}

@inproceedings{abramovich2024visfocus,
  title={Visfocus: Prompt-guided vision encoders for ocr-free dense document understanding},
  author={Abramovich, Ofir and Nayman, Niv and Fogel, Sharon and Lavi, Inbal and Litman, Ron and Tsiper, Shahar and Tichauer, Royee and Appalaraju, Srikar and Mazor, Shai and Manmatha, R},
  booktitle={European Conference on Computer Vision},
  pages={241--259},
  year={2024},
  organization={Springer}
}

@article{mao2026language,
  title={Language-Instructed Vision Embeddings for Controllable and Generalizable Perception},
  author={Mao, Chengzhi and Lin, Xudong and Chu, Wen-Sheng},
  journal={arXiv preprint arXiv:2606.19584},
  year={2026}
}

@article{liu2026igvlm,
  title={iGVLM: Dynamic Instruction-Guided Vision Encoding for Question-Aware Multimodal Understanding},
  author={Liu, Hanpeng and Li, Yaqian and Wang, Zidan and Zhang, Shuoxi and Bo, Zihao and Takezoe, Rinyoichi and Long, Kaiwen and He, Kun},
  journal={arXiv preprint arXiv:2603.02748},
  year={2026}
}

@inproceedings{ganz2024question,
  title={Question aware vision transformer for multimodal reasoning},
  author={Ganz, Roy and Kittenplon, Yair and Aberdam, Aviad and Ben Avraham, Elad and Nuriel, Oren and Mazor, Shai and Litman, Ron},
  booktitle={Proceedings of the IEEE/CVF conference on computer vision and pattern recognition},
  pages={13861--13871},
  year={2024}
}

@inproceedings{perez2018film,
  title={Film: Visual reasoning with a general conditioning layer},
  author={Perez, Ethan and Strub, Florian and De Vries, Harm and Dumoulin, Vincent and Courville, Aaron},
  booktitle={Proceedings of the AAAI conference on artificial intelligence},
  volume={32},
  number={1},
  year={2018}
}

@article{de2017modulating,
  title={Modulating early visual processing by language},
  author={De Vries, Harm and Strub, Florian and Mary, J{\'e}r{\'e}mie and Larochelle, Hugo and Pietquin, Olivier and Courville, Aaron},
  journal={Advances in neural information processing systems},
  volume={30},
  year={2017}
}

@article{raghu2021vision,
  title={Do vision transformers see like convolutional neural networks?},
  author={Raghu, Maithra and Unterthiner, Thomas and Kornblith, Simon and Zhang, Chiyuan and Dosovitskiy, Alexey},
  journal={Advances in neural information processing systems},
  volume={34},
  pages={12116--12128},
  year={2021}
}

@article{bolya2026perception,
  title={Perception encoder: The best visual embeddings are not at the output of the network},
  author={Bolya, Daniel and Huang, Po-Yao and Sun, Peize and Cho, Jang Hyun and Madotto, Andrea and Wei, Chen and Ma, Tengyu and Zhi, Jiale and Rajasegaran, Jathushan and Bangalath, Hanoona and others},
  journal={Advances in Neural Information Processing Systems},
  volume={38},
  pages={60884--60937},
  year={2026}
}

@inproceedings{tong2024eyes,
  title={Eyes wide shut? exploring the visual shortcomings of multimodal llms},
  author={Tong, Shengbang and Liu, Zhuang and Zhai, Yuexiang and Ma, Yi and LeCun, Yann and Xie, Saining},
  booktitle={2024 IEEE/CVF Conference on Computer Vision and Pattern Recognition (CVPR)},
  pages={9568--9578},
  year={2024},
  organization={IEEE}
}

@inproceedings{kornblith2019similarity,
  title={Similarity of neural network representations revisited},
  author={Kornblith, Simon and Norouzi, Mohammad and Lee, Honglak and Hinton, Geoffrey},
  booktitle={International conference on machine learning},
  pages={3519--3529},
  year={2019},
  organization={PMlR}
}

@article{huh2024platonic,
  title={The platonic representation hypothesis},
  author={Huh, Minyoung and Cheung, Brian and Wang, Tongzhou and Isola, Phillip},
  journal={arXiv preprint arXiv:2405.07987},
  year={2024}
}

@inproceedings{zhai2023sigmoid,
  title={Sigmoid loss for language image pre-training},
  author={Zhai, Xiaohua and Mustafa, Basil and Kolesnikov, Alexander and Beyer, Lucas},
  booktitle={2023 IEEE/CVF International Conference on Computer Vision (ICCV)},
  pages={11941--11952},
  year={2023},
  organization={IEEE}
}

@article{bai2025qwen3,
  title={Qwen3-vl technical report},
  author={Bai, Shuai and Cai, Yuxuan and Chen, Ruizhe and Chen, Keqin and Chen, Xionghui and Cheng, Zesen and Deng, Lianghao and Ding, Wei and Gao, Chang and Ge, Chunjiang and others},
  journal={arXiv preprint arXiv:2511.21631},
  year={2025}
}
}

\clearpage
\appendix

\section{Phase-Aware Recurrent Window Selection}
\label{sec:app_phase_window}

\subsection{Layer Similarity and Visual Blocks}
\label{sec:app_layer_similarity}

We analyze the internal block structure of each vision tower with layer-wise centered linear CKA \citep{kornblith2019similarity}. For a model with $L$ visual layers, we collect token representations from each layer on a small shared image set and compute an $L\times L$ similarity matrix. Bright diagonal regions indicate contiguous phases of layers with aligned representations. We refer to each coherent phase as a visual \emph{block}, following the block-recurrent view that Transformer depth can often be described by a small number of reusable computational blocks \citep{dehghani2018universal,jacobs2026block}.

This analysis is used as a structural guide for recurrent window placement. A window inside one block applies a visually consistent transformation across loops. A window that spans several blocks mixes different stages of visual processing, such as local appearance extraction and semantic integration. The main paper therefore uses middle-to-late blocks as the default search region and validates the choice with window sweeps.

\subsection{CKA Maps Across Backbones}
\label{sec:app_cka_maps}

\begin{figure*}[t]
    \centering
    \begin{subfigure}[t]{0.32\linewidth}
        \centering
        \includegraphics[width=\linewidth]{figures/layersim_q25.pdf}
        \caption{Qwen2.5-VL-3B}
    \end{subfigure}
    \hfill
    \begin{subfigure}[t]{0.32\linewidth}
        \centering
        \includegraphics[width=\linewidth]{figures/layersim_q3.pdf}
        \caption{Qwen3-VL-4B}
    \end{subfigure}
    \hfill
    \begin{subfigure}[t]{0.32\linewidth}
        \centering
        \includegraphics[width=\linewidth]{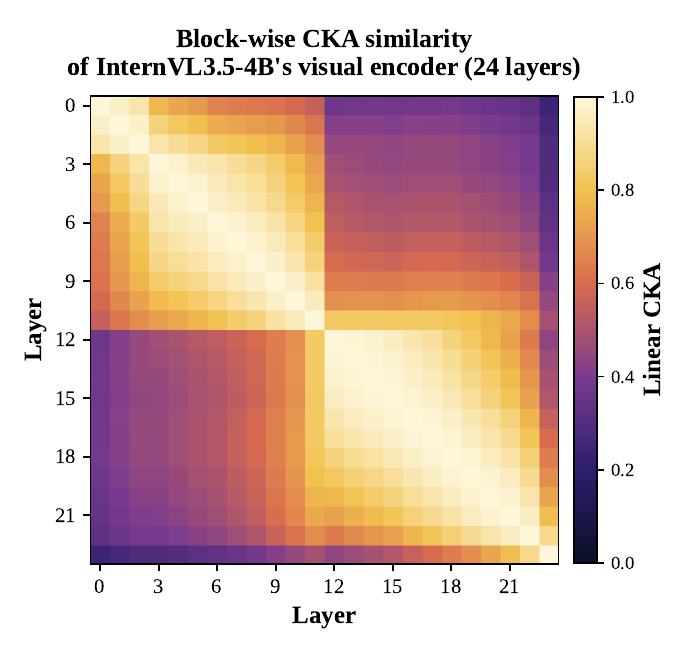}
        \caption{InternVL3.5-4B}
    \end{subfigure}

    \vspace{3pt}
    \begin{subfigure}[t]{0.32\linewidth}
        \centering
        \includegraphics[width=\linewidth]{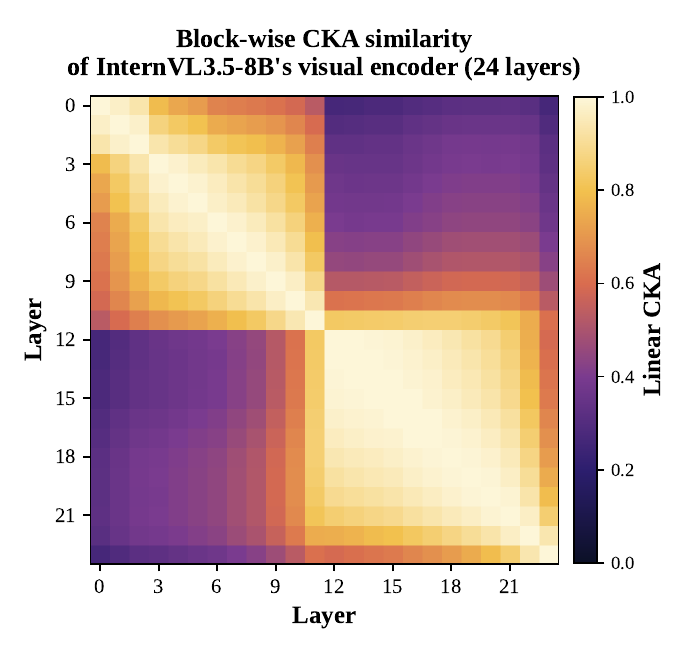}
        \caption{InternVL3.5-8B}
    \end{subfigure}
    \hfill
    \begin{subfigure}[t]{0.32\linewidth}
        \centering
        \includegraphics[width=\linewidth]{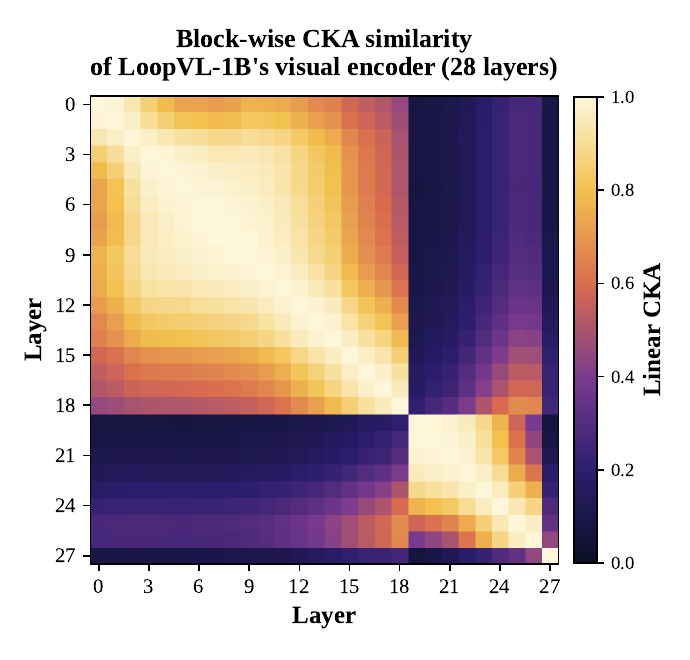}
        \caption{LoopVL-1B}
    \end{subfigure}
    \hfill
    \begin{subfigure}[t]{0.32\linewidth}
        \centering
        \includegraphics[width=\linewidth]{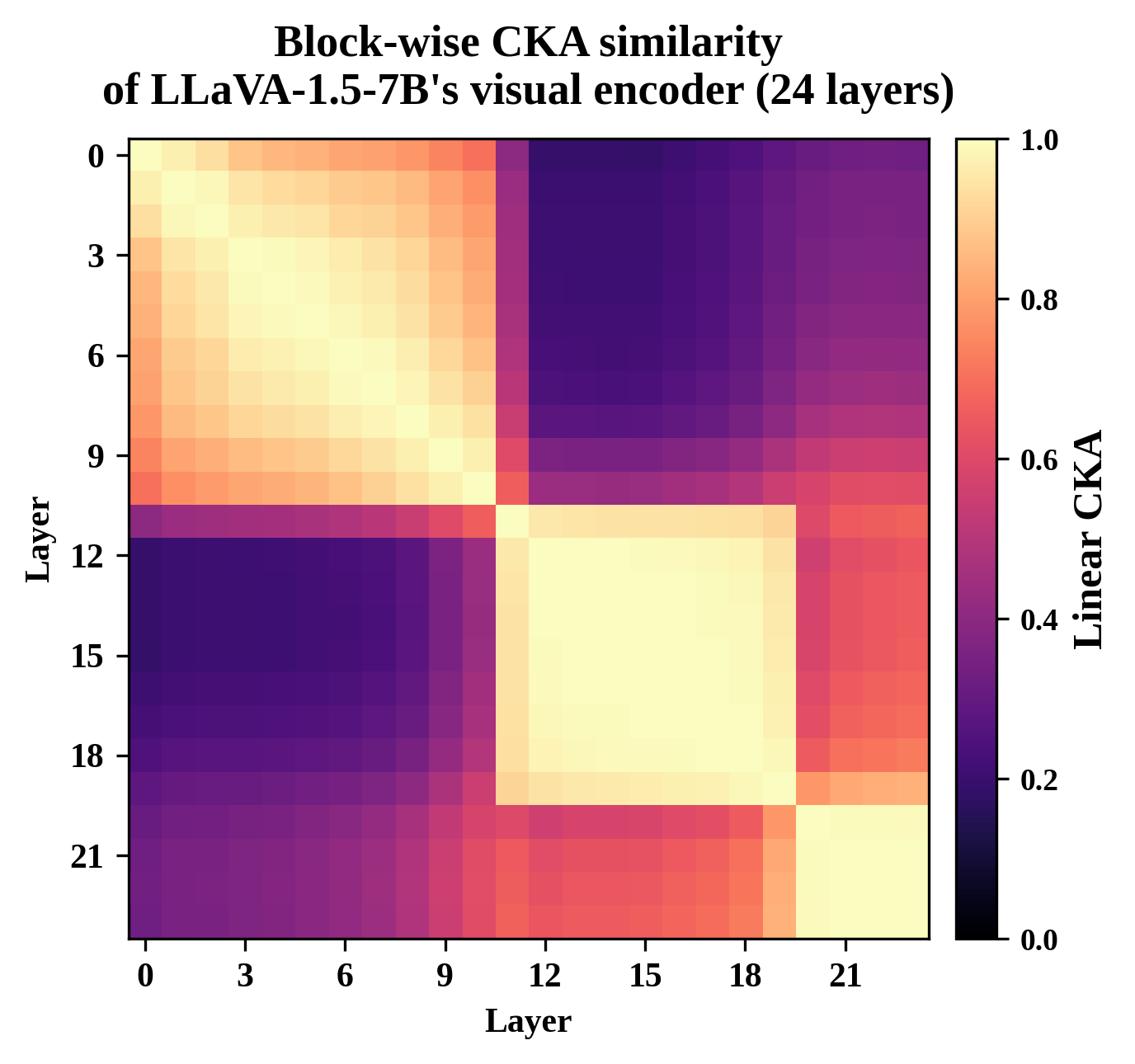}
        \caption{LLaVA-1.5-7B}
    \end{subfigure}
    \caption{Layer-wise linear CKA maps for six visual towers. Across backbones, the encoder depth is organized into a small number of coherent visual blocks.}
    \label{fig:app_cka_all}
\end{figure*}

\Cref{fig:app_cka_all} shows the full CKA maps for the six visual towers analyzed in this work, covering Qwen2.5-VL \citep{bai2025qwen25vltechnicalreport}, Qwen3-VL \citep{bai2025qwen3}, InternVL3.5 \citep{wang2025internvl3}, LoopVL \citep{qian2026loopvl}, and LLaVA \citep{liu2024improved} backbones. Across these models, the matrices consistently exhibit a small number of bright diagonal blocks, suggesting that pretrained visual encoders organize depth into coherent computational phases. \Cref{tab:cka_boundaries} summarizes the corresponding CKA blocks, whose locations vary by architecture and motivate phase-aware window selection for each backbone.

\subsection{Window Sweep Protocol}
\label{sec:app_window_sweep}

For each backbone, we sweep recurrent windows around the middle and late encoder regions and compare their downstream performance under matched training and inference settings. The CKA maps serve as a guide for selecting candidate windows; the final claim is based on the window sweep. In practice, windows that remain inside a coherent middle-to-late block give more stable gains, while very early windows and windows that cross strong block boundaries are less reliable. This supports the use of block-consistent recurrent visual encoding in the main method.

\begin{table}[t]
    \centering
    \caption{CKA blocks for the visual towers analyzed in this work. Layer ranges are zero-indexed and interpreted from the CKA maps; singleton ranges denote isolated readout-style layers. Bold layer ranges indicate the blocks selected in our experiments for recurrent processing in each backbone.}
    \label{tab:cka_boundaries}
    \small
    \begin{tabular}{lcc}
        \toprule
        Backbone & Layers & CKA blocks \\
        \midrule
        Qwen2.5-VL-3B & 32 & 0--14, 15--16, \textbf{17--30}, 31 \\
        Qwen3-VL-4B & 24 & 0--9, \textbf{10--22}, 23 \\
        InternVL3.5-4B & 24 & 0--11, \textbf{12--22}, 23 \\
        InternVL3.5-8B & 24 & 0--11, 12--22, 23 \\
        LoopVL-1B & 28 & 0--18, 19--26, 27 \\
        LLaVA-1.5-7B & 24 & 0--10, \textbf{11--19}, 20--23 \\
        \bottomrule
    \end{tabular}
\end{table}

\section{More Case Studies}
\label{sec:app_case_studies}

This section collects qualitative case studies. \Cref{sec:app_case_baby} presents two open-ended BabyVision questions answered correctly by InternVL3.5-4B, where the free-form responses can be verified step by step. \Cref{sec:app_case_refine} presents failure-to-success comparisons between the base VLM and LoopLook, visualizing how the recurrent visual features progressively refine attention onto the evidence region.

\subsection{BabyVision Open-Ended Cases}
\label{sec:app_case_baby}

\paragraph{Case A: Counting stacked cubes.}

\begin{figure*}[t]
    \centering
    \begin{minipage}{0.94\textwidth}
        \centering
        \includegraphics[width=0.40\linewidth]{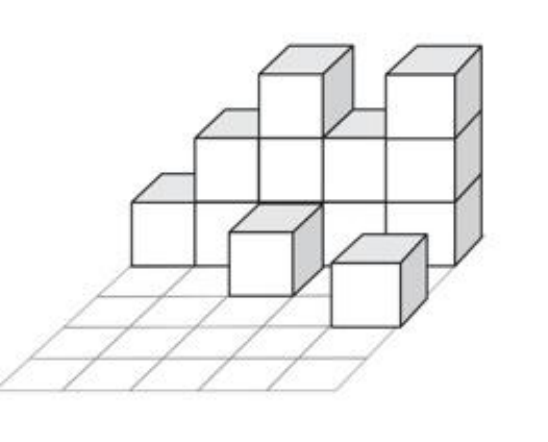}\\[8pt]
        \begin{promptbox}[Question]
        The image shows a three-dimensional shape made of stacked cubes. None of the cubes are suspended in the air. How many cubes are there at least?
        \end{promptbox}

        \begin{promptbox}[Response (correct; ground truth: 13)]
        To determine the minimum number of cubes in the given three-dimensional shape, we need to carefully analyze the structure and ensure that we account for all visible and hidden cubes. The shape has a base layer with several cubes; on top of some of these base cubes, there are additional cubes. The base layer has a total of 7 cubes, and there are 6 cubes in the second and third layer. Therefore, the total number of cubes is 7 + 6 = 13. Thus, the minimum number of cubes in the shape is \textbackslash boxed\{13\}.
        \end{promptbox}
    \end{minipage}
    \caption{A correct LoopLook open-ended counting case. The model decomposes the stacked-cube structure into a base layer and upper layers, verifies that every upper cube is supported, and reaches the minimum count of 13.}
    \label{fig:app_case_baby_counting}
\end{figure*}

An example is shown in \cref{fig:app_case_baby_counting}.

\paragraph{Case B: Subtracting two groups of planets.}

\begin{figure*}[t]
    \centering
    \begin{minipage}{0.94\textwidth}
        \centering
        \includegraphics[width=0.86\linewidth]{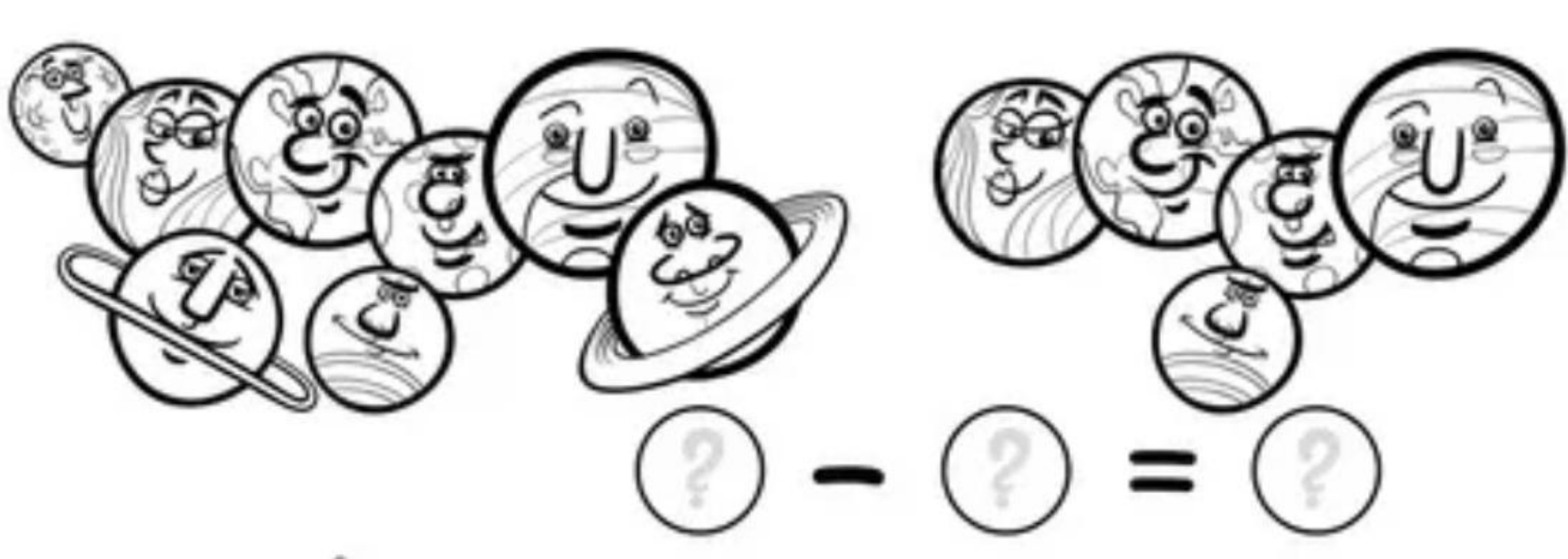}\\[8pt]
        \begin{promptbox}[Question]
        Based on the number of cartoon planets in the left and right groups, list the subtraction equation and calculate the result, and write it in the answer in the following format, for example: 3-1=2.
        \end{promptbox}

        \begin{promptbox}[Response (correct; ground truth: 8-5=3)]
        To solve the problem, we need to count the number of cartoon planets in each group. On the left side, there are 8 cartoon planets. On the right side, there are 5 cartoon planets. The subtraction equation is: 8 - 5 = 3. So, the answer is: \textbackslash boxed\{8 - 5 = 3\}.
        \end{promptbox}
    \end{minipage}
    \caption{A correct LoopLook open-ended subtraction case. The model counts the two groups of cartoon planets separately and forms the correct subtraction equation.}
    \label{fig:app_case_baby_sub}
\end{figure*}

Another example is shown in \cref{fig:app_case_baby_sub}.
\begin{figure*}[t]
    \centering
    \begin{subfigure}[t]{0.95\textwidth}
        \centering
        \includegraphics[width=\linewidth]{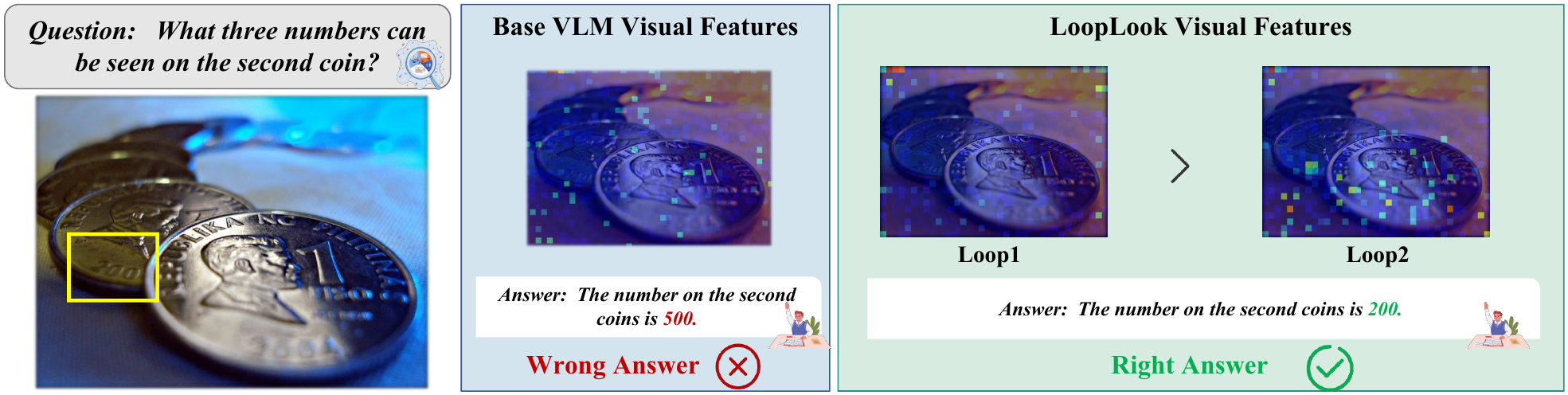}
        \caption{Reading the numbers on a coin. The base attention scatters over the coin face and misreads the digits as ``500''; the looped features sharpen onto the stamped region and recover all three digits (``200'').}
        \label{fig:app_case_refine-a}
    \end{subfigure}

    \vspace{8pt}
    \begin{subfigure}[t]{0.95\textwidth}
        \centering
        \includegraphics[width=\linewidth]{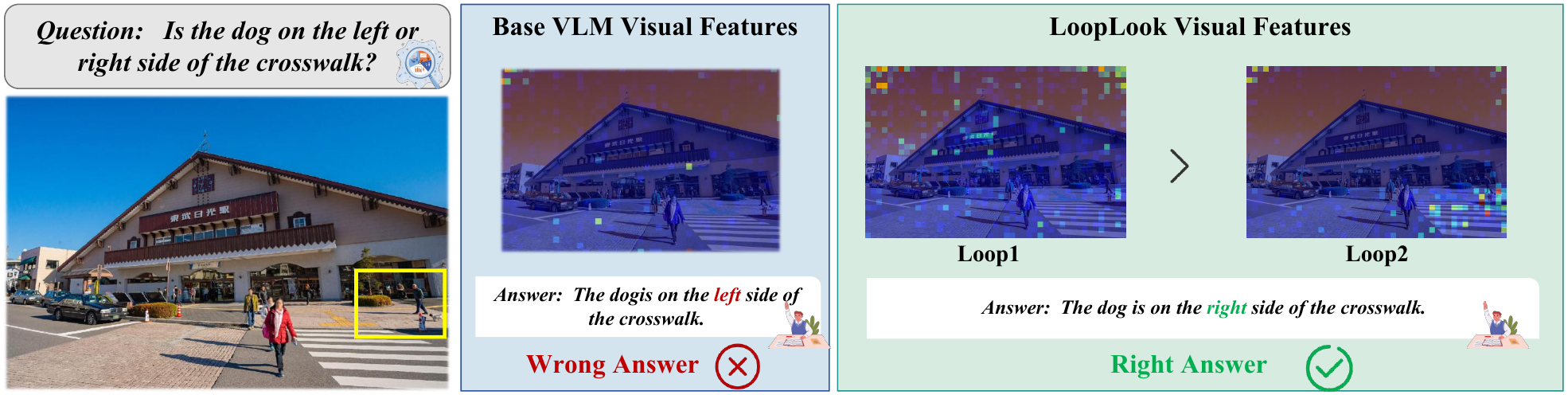}
        \caption{Localizing a dog relative to a crosswalk. The base attention locks onto the storefront and answers ``left''; the looped features re-anchor onto the dog beside the crossing and correct the answer to ``right''.}
        \label{fig:app_case_refine-b}
    \end{subfigure}

    \vspace{8pt}
    \begin{subfigure}[t]{0.95\textwidth}
        \centering
        \includegraphics[width=\linewidth]{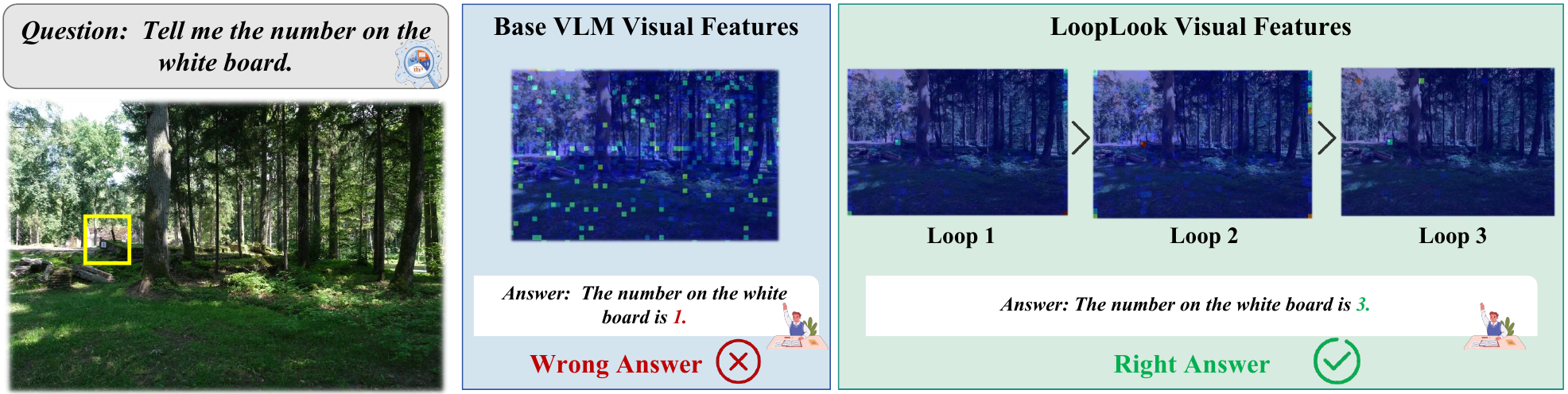}
        \caption{Reading a number on a whiteboard deep inside a forest scene. The base attention diffuses over the tree canopy; the looped features refine onto the small whiteboard and read the digit correctly.}
        \label{fig:app_case_refine-c}
    \end{subfigure}

    \vspace{8pt}
    \begin{subfigure}[t]{0.95\textwidth}
        \centering
        \includegraphics[width=\linewidth]{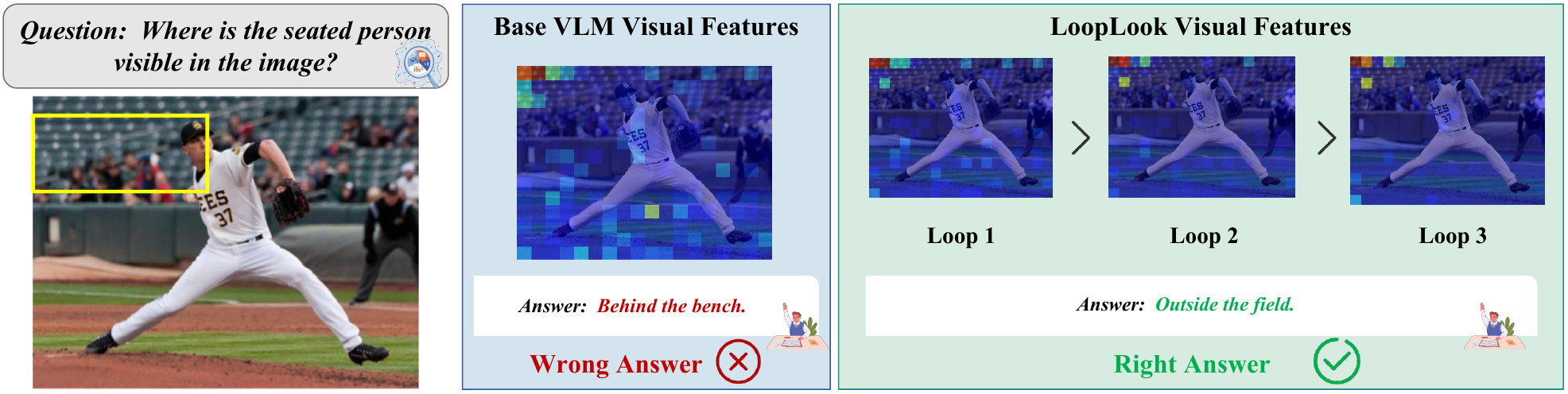}
        \caption{Finding a seated person in a baseball scene. The base attention follows the pitcher in the foreground; the looped features expand toward the spectator area outside the field and localize the seated person correctly.}
        \label{fig:app_case_refine-d}
    \end{subfigure}
    \caption{Failure-to-success comparisons between the base VLM and LoopLook. Yellow boxes mark the queried evidence region.}
    \label{fig:app_case_refine}
\end{figure*}

\begin{figure*}[t]
    \centering
    \begin{subfigure}[t]{0.85\textwidth}
        \centering
        \includegraphics[width=\linewidth]{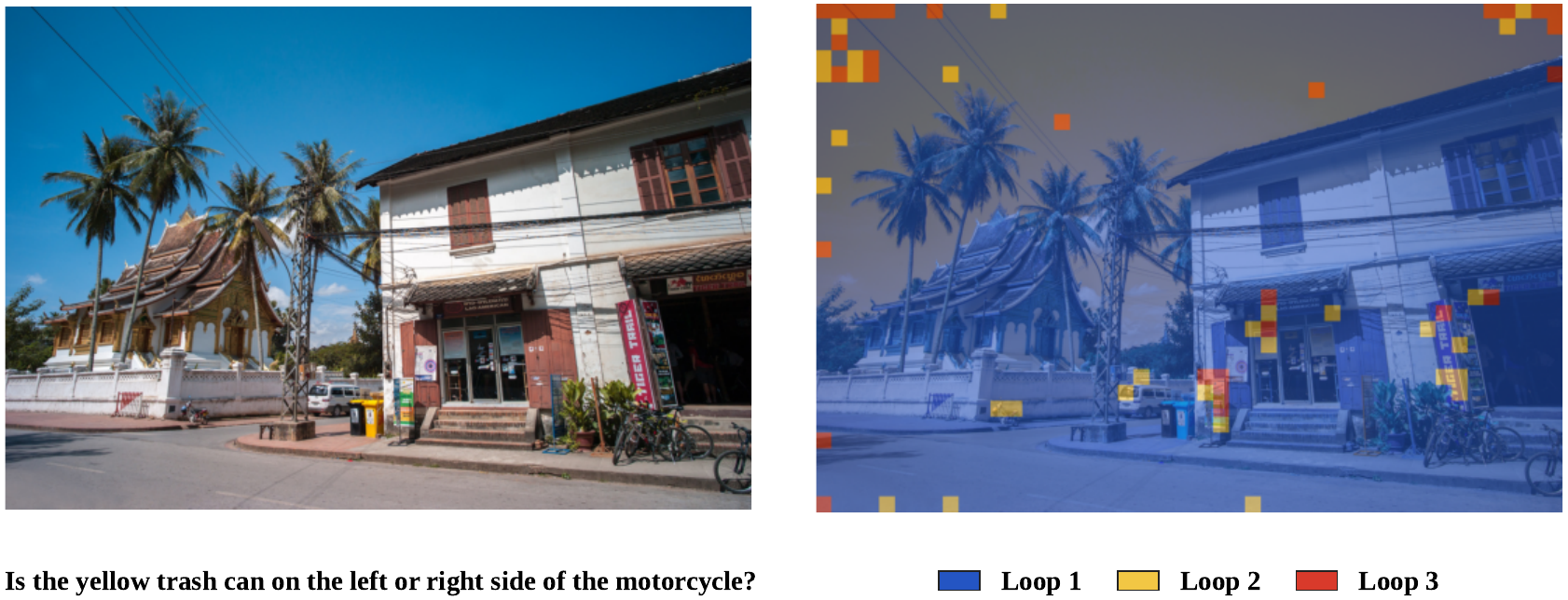}
        \caption{Relative-position case.}
        \label{fig:app_case_tiers-a}
    \end{subfigure}

    \vspace{8pt}
    \begin{subfigure}[t]{0.85\textwidth}
        \centering
        \includegraphics[width=\linewidth]{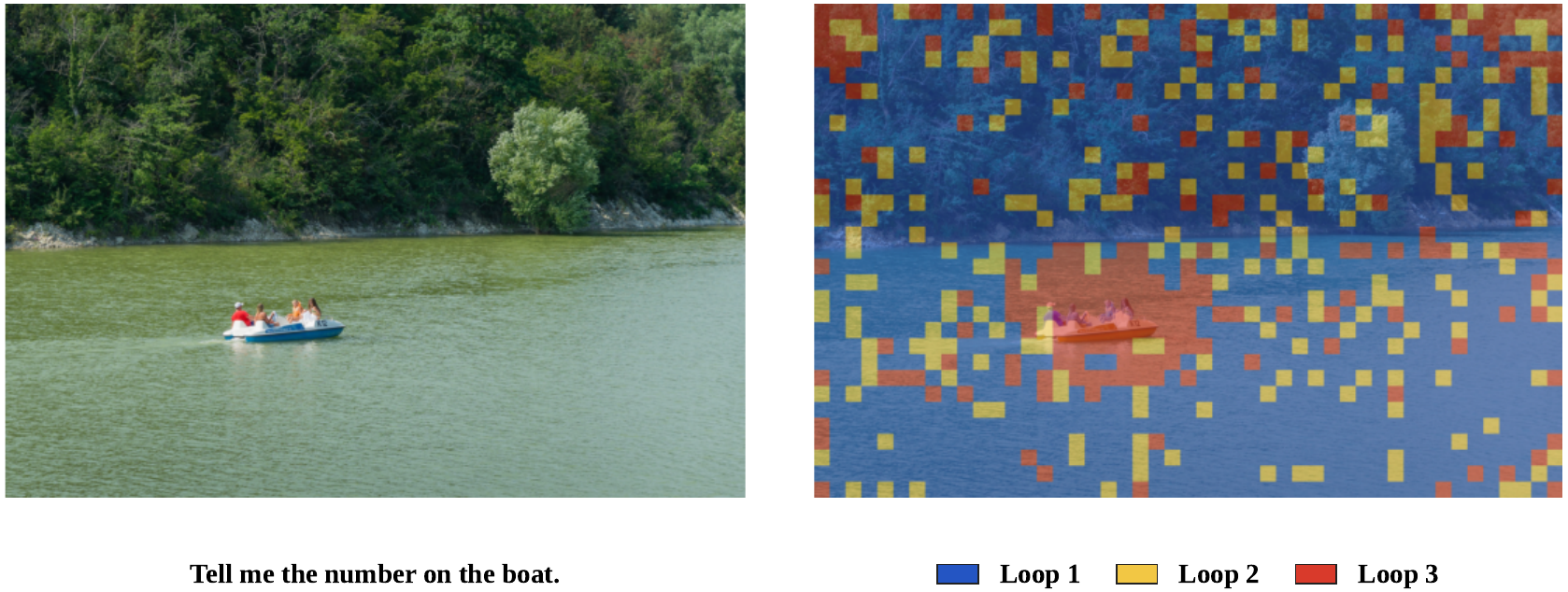}
        \caption{Fine-grained OCR case.}
        \label{fig:app_case_tiers-b}
    \end{subfigure}

    \vspace{8pt}
    \begin{subfigure}[t]{0.85\textwidth}
        \centering
        \includegraphics[width=\linewidth]{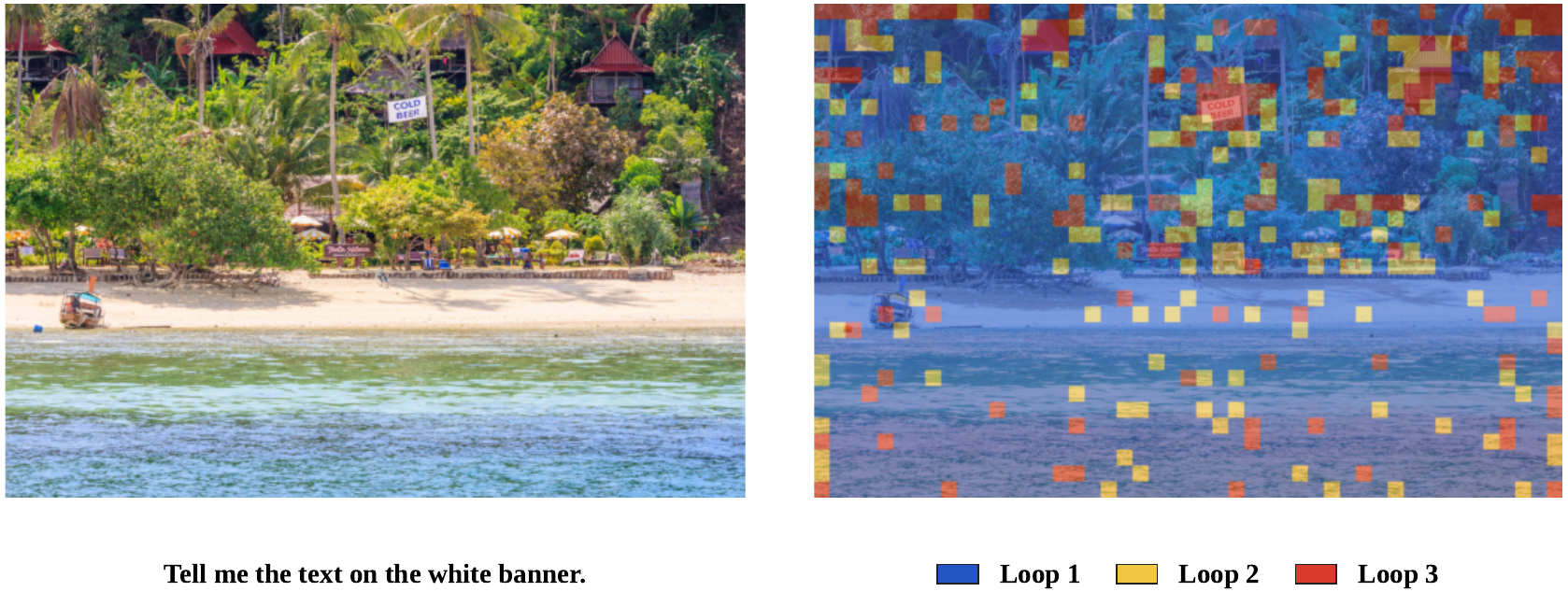}
        \caption{Text-reading case.}
        \label{fig:app_case_tiers-c}
    \end{subfigure}
    \caption{Per-token dominant-loop tier maps on three additional cases. Blue/yellow/red denote tokens whose evidence is dominated by Loop 1/2/3 respectively.}
    \label{fig:app_case_tiers}
\end{figure*}

\subsection{Attention Refinement Cases}
\label{sec:app_case_refine}

\Cref{fig:app_case_refine} shows representative cases in which the base VLM answers incorrectly while LoopLook succeeds. Each panel visualizes the attention distribution of the base VLM and of LoopLook at every recurrent step. In all cases, the base attention spreads over background regions, whereas the looped features progressively concentrate onto the evidence region. The number of loops actually used varies with the difficulty of localization (two loops for \cref{fig:app_case_refine-a,fig:app_case_refine-b}, three loops for \cref{fig:app_case_refine-c,fig:app_case_refine-d}).

\subsection{Per-Token Loop Tier Visualization}
\label{sec:app_case_tiers}

\Cref{fig:app_case_tiers} further visualizes the 
token-wise Mixture-of-Looks decision by assigning each
image token to the loop state with the largest routing weight.

\section{Implementation Details}
\label{sec:app_impl}

\subsection{Backbones and Trainable Parameters}
\label{sec:app_impl_backbones}
We implement LoopLook on four pretrained VLM backbones: Qwen2.5-VL-3B \citep{bai2025qwen25vltechnicalreport}, Qwen3-VL-4B \citep{bai2025qwen3}, InternVL3.5-4B \citep{wang2025internvl3}, and LLaVA-1.5-7B \citep{liu2024improved}. For each backbone, the recurrent window is placed in the middle-to-late visual block selected by the phase analysis in \cref{sec:app_phase_window}. During training, we fine-tune the pretrained vision encoder using LoRA \citep{hu2021lora} with a rank of \(r=8\), a scaling factor of \(\alpha=16\), and a dropout rate of \(0.05\). The resulting trainable budgets are 14.9M, 10.6M, 10.1M, and 11.1M parameters for Qwen2.5-VL-3B, Qwen3-VL-4B, InternVL3.5-4B, and LLaVA-1.5-7B respectively, all below 0.4\% of each backbone's total parameters (see \cref{tab:app_trainable}). All parameters of the encoder-side LoopLook modules are fully trainable, including the recurrent perception components, adaptive feature integration modules, and exit mechanisms. The language model is kept frozen throughout training. This configuration combines parameter-efficient adaptation of the pretrained vision encoder with full optimization of the newly introduced LoopLook modules. This setting keeps the original language generation interface unchanged and concentrates adaptation on visual perception.
\begin{table*}[t]
    \centering
    \caption{Trainable-parameter composition of LoopLook (millions).}
    \label{tab:app_trainable}
    \scriptsize
    \setlength{\tabcolsep}{4pt}
    \renewcommand{\arraystretch}{1.1}
    \resizebox{0.8\textwidth}{!}{%
    \begin{tabular}{lcccc}
        \toprule
        Component & Qwen2.5-VL-3B & Qwen3-VL-4B & InternVL3.5-4B & LLaVA-1.5-7B \\
        \midrule
        LoRA (vision tower, $r{=}8$) & 4.37 & 3.15 & 3.15 & 3.15 \\
        Task projection ($d_l{\to}d$) & 2.62 & 2.62 & 2.62 & 4.20 \\
        $M_z$ & 0.44 & 0.28 & 0.28 & 0.28 \\
        $M_\ell$ (all recurrent blocks) & 5.77 & 3.43 & 2.91 & 2.38 \\
        Loop-step embeddings $\mathbf{e}_r$ & 0.001 & 0.001 & 0.001 & 0.001 \\
        Exit head & 0.87 & 0.56 & 0.56 & 0.56 \\
        Mixture-of-Looks router & 0.87 & 0.56 & 0.56 & 0.56 \\
        \midrule
        \textbf{Total (\% of backbone)} & \textbf{14.9} (0.40\%) & \textbf{10.6} (0.25\%) & \textbf{10.1} (0.24\%) & \textbf{11.1} (0.16\%) \\
        \bottomrule
    \end{tabular}%
    }
\end{table*}
\subsection{LoopLook Hyperparameters}
\label{sec:app_impl_hparams}
For all backbones, we set the maximum loop depth to $R_{\max}=4$ and use an exit threshold of $0.4$ at inference time. The modulation state $\mathbf{z}^{r}$ lies in a compact subspace of dimension $d/8$, and the loop-step embedding $\mathbf{e}_{r}$ shares this dimension. For each recurrent block $\ell$, $M_{\ell}$ is a single linear map from this state to the $2d$ affine coefficients $(\gamma_{\ell}^{r}, \beta_{\ell}^{r})$. We use mean pooling over visual tokens to obtain the visual summary in both the modulation module and the exit gate. The token-wise Mixture-of-Looks router is implemented as a two-layer MLP with hidden size $d/4$; the exit head uses the same hidden width and outputs a scalar sigmoid score after each loop. 

\subsection{Training Setup}
\label{sec:app_impl_training}
We train LoopLook with 100K Visual CoT \citep{shao2024visual} samples for one epoch using a global batch size of $8$. We use AdamW with learning rate $3\times10^{-5}$, $\beta=(0.9,0.95)$, and weight decay $10^{-2}$. The depth penalty uses a linear ramp with $\lambda_d^{\star}=0.0005$, $t_w=0.6T_{\mathrm{train}}$, and $t_r=0.4T_{\mathrm{train}}$, where $T_{\mathrm{train}}$ denotes the total number of training steps. Experiments are trained in parallel on eight RTX 5090 GPUs. We apply global gradient clipping with norm $1.0$.
\subsection{Evaluation Settings}
\label{sec:app_impl_eval}
We evaluate all models with VLM-EvalKit\footnote{\url{https://github.com/open-compass/VLMEvalKit}}. For the Base and CoT baselines, we use greedy decoding with \texttt{max\_new\_tokens=512}. For Self-Consistency, we sample $5$ CoT responses with temperature $0.5$ and \texttt{max\_new\_tokens=512}, then predict by majority vote. All reported LoopLook results use the same benchmark prompts and answer extraction protocol as the corresponding backbone evaluations.

\subsection{Efficiency Measurement}
\label{sec:app_efficiency}

We evaluate the inference efficiency of Base and LoopLook on a single NVIDIA GeForce RTX 5090 GPU (32 GB). We randomly sample 200 examples from the downstream benchmarks and use the same examples and decoding configuration for both methods. For each model, we perform five warm-up forward passes before timing, excluding warm-up runs from the measurements. We evaluate Base and LoopLook in an interleaved order to mitigate thermal drift and order-dependent measurement bias. CUDA operations are synchronized at timing boundaries to ensure accurate wall-clock measurements.

For each input resolution, we generate 64 tokens per example using identical decoding settings. We report three metrics: (1) \textit{time to first token} (TTFT), measured from the start of the timed inference request to the return of the first generated token; (2) \textit{decoding latency}, measured as the average wall-clock time per generated token during autoregressive decoding; and (3) \textit{end-to-end} (E2E) latency, measured from the start of the timed inference request to the completion of 64-token generation. We repeat the measurements and report the mean latency across runs.

All measurements are conducted using Python 3.13.5, PyTorch 2.14.0+cu130, CUDA 13.0, cuDNN 9.24, and Transformers 5.16.1, with NVIDIA driver version 580.95.05.

\section{Ablation Protocol}
\label{sec:app_ablation_protocol}
\subsection{Shared Training and Evaluation Settings}
\label{sec:app_ablation_settings}

We evaluate controls on Qwen2.5-VL-3B targeting additional fine-tuning (Param-Matched FT and FLOPs-Matched FT), recurrent visual processing, task conditioning, token-wise readout, and adaptive stopping. The five training-based controls are trained independently from the same pretrained backbone initialization, while the fixed-depth \textit{w/o adaptive exit} variants reuse the full LoopLook checkpoint and change only the inference policy. Task-conditioned variants derive the visual task state from the instruction. All configurations follow the training and evaluation protocols in \cref{sec:app_impl}, including prompts, answer extraction, and benchmark scoring. We report MMStar, V$^\ast$, and HallusionBench, with Avg. defined as their arithmetic mean.

\subsection{Param- and FLOPs-Matched Fine-Tuning}
\label{sec:app_ablation_matched_ft}

\paragraph{Param-Matched FT.}
This control measures the gains from additional visual-side fine-tuning under a comparable trainable-parameter budget. We retain the original one-pass visual path,
\begin{equation}
    \mathbf V=P_\psi(E_\theta(\mathbf I)),
\end{equation}
including its original feature selection and projector interface. No recurrent window execution, task-conditioned modulation, exit gate, or Mixture-of-Looks module is added. The baseline is trained on the same image--instruction--response examples using the autoregressive answer loss, without a depth penalty.

To match the full model's trainable-parameter budget, the baseline uses visual LoRA with rank $28$ and scaling $56$, preserving $\alpha/r=2$. LoRA injection locations and dropout remain unchanged. The baseline has approximately $15.3$M trainable parameters, compared with $14.9$M for the full LoopLook model.

\paragraph{FLOPs-Matched FT.}
Param-matched training does not guarantee equal computational budgets: LoopLook executes its recurrent window $R_{\max}=4$ times per sample, while Param-Matched FT processes the vision tower only once. We estimate training FLOPs using $4NT$ per linear layer with $N$ parameters and $T$ tokens, accounting for both forward and backward passes. Since the language model is frozen, backward computation only requires input gradients. We distinguish visual patches $T_p$ from merged visual tokens $T_m=T_p/4$ processed by the language model. For Qwen2.5-VL-3B, with $N_{\mathrm{LLM}}=3.09$B, $N_{\mathrm{ViT}}=0.49$B across 32 layers, and $N_{\mathrm{W}}=214$M across the 14-layer window (blocks 17--30), we use $T_p\approx1531$, $T_m\approx383$, and $T_t\approx250$ at $300$K pixels:
\begin{equation}
\footnotesize
\begin{gathered}
F_{\mathrm{base}}
=4N_{\mathrm{LLM}}(T_m{+}T_t)+4N_{\mathrm{ViT}}T_p
\approx 10.8\ \mathrm{TF},\\
F_{\mathrm{LoopLook}}
=F_{\mathrm{base}}+(R_{\max}{-}1)\,4N_{\mathrm{W}}T_p
\approx 14.8\ \mathrm{TF}.
\end{gathered}
\label{eq:app_flops}
\end{equation}
Only three window executions add extra FLOPs, as the first is included in the full tower pass. The trainable LoopLook modules ($M_z$, $M_\ell$, exit head, and router; approximately $10.5$M parameters) add less than $0.5\%$ to the estimated budget. Under the same estimate, the LoopLook-to-baseline FLOPs ratio is $\approx\!1.36\times$ for Qwen2.5-VL-3B, $\approx\!1.25\times$ for Qwen3-VL-4B and InternVL3.5-4B, and $\approx\!1.03\times$ for LLaVA-1.5-7B, whose frozen language model dominates computation. To ensure a fair comparison, we train the same Param-Matched FT baseline for 1.5× as many optimization steps under the native cosine schedule, giving it a slightly larger training compute budget than LoopLook's estimated overhead requires across all evaluated backbones.

\subsection{w/o Recurrence}
\label{sec:app_ablation_single_look}

We set $R_{\max}=1$ throughout training, so the visual window executes only once. The task representation, initial visual state, modulation networks, and first-step embedding are retained:
\begin{equation}
    \mathbf H^1=F_\theta^{s:e}(\mathbf H^0;\mathbf x,0),
    \qquad
    \tilde{\mathbf H}=\mathbf H^1.
\end{equation}
The update uses the same block-wise adaptive modulation as the full model, followed by the standard readout tail and projector. Visual LoRA and task-conditioned modulation remain trainable. This control compares a single task-conditioned visual update with recurrent refinement.

\subsection{w/o Task Conditioning}
\label{sec:app_ablation_no_task}

We replace the visual-side task vector with a fixed zero vector, $\mathbf q_{\mathrm{visual}}=\mathbf 0$, while preserving the original interfaces and dimensions. This replacement is applied to the modulation network $M_z$, exit head $h_\omega$, and readout router $g_\eta$. The language model still receives the original instruction, and the target response and answer supervision remain unchanged.

Visual-state conditioning, loop-step embeddings, the modulation architecture, adaptive exit, token-wise MoL, and the original training objective are retained. The model can adapt its visual computation to the image without question guidance. Comparing this variant with the full model measures the contribution of task conditioning across the three visual-side branches.

\subsection{w/o Token-Wise Fusion}
\label{sec:app_ablation_global_mol}

The \textit{w/o token-wise fusion} variant replaces token-specific fusion weights with a shared set of weights for each image. We retain the original scoring network and aggregate its token-wise logits:
\begin{equation}
\begin{aligned}
    \ell_{i,r}&=g_\eta(\mathbf h_i^r,\mathbf q,\mathbf p_i,r),\\
    \bar\ell_r&=\frac{1}{N}\sum_{i=1}^{N}\ell_{i,r}.
\end{aligned}
\end{equation}
The averaged logits and stopping prior define a shared distribution over the available loops. With $K=R_{\max}$ during training and $K=\hat R$ at inference, its weights are
\begin{equation}
    \beta_r
    =\operatorname{softmax}_{r\le K}\!\left(
    \bar\ell_r+\log(\pi_r+\epsilon)\right),
\end{equation}
and the same weights are applied to every token:
\begin{equation}
    \tilde{\mathbf h}_i
    =\sum_{r=1}^{K}
    \beta_r\mathbf h_i^r.
\end{equation}
The loop distribution varies across images but is shared across tokens within each image. The scoring network and parameter count are unchanged.

\subsection{Fixed-Depth Execution (w/o Adaptive Exit)}
\label{sec:app_ablation_fixed_depth}

We use the full LoopLook checkpoint trained with $R_{\max}=4$ and set a fixed inference depth $K\in\{1,2,3,4\}$. For each setting, every sample executes exactly $K$ loops, and Mixture-of-Looks combines the resulting states. The exit head supplies the stopping prior with the original $R_{\max}=4$ setting:
\begin{equation}
\begin{gathered}
    \alpha_{i,r}
    =\operatorname{softmax}_{r=1,\ldots,K}
    \!\left(g_{i,r}+\log(\pi_r+\epsilon)\right),\\
    \tilde{\mathbf h}_i
    =\sum_{r=1}^{K}\alpha_{i,r}\mathbf h_i^r.
\end{gathered}
\end{equation}
The adaptive variant uses the same checkpoint and readout, with $K=\hat R$ selected by the exit gate.

\section{LoopLook Training and Inference Algorithm}
\label{sec:app_algorithm}

Algorithm~\ref{alg:looplook} summarizes task-conditioned recurrence, adaptive stopping, and Mixture-of-Looks readout. Training unrolls all recurrent steps, while inference computes the prefix selected by the exit gate. In both modes, the router fuses the available states using token-specific scores and the stopping prior, then passes the fused representation through the tail and projector. Training applies the answer loss and depth penalty to optimize the visual-side parameters.

\begin{algorithm*}[t]
\small
\caption{Training and inference workflow of LoopLook}
\label{alg:looplook}
\Input{Image $\mathbf{I}$, instruction $\mathbf{x}$, mode $m\in\{\mathrm{train},\mathrm{infer}\}$; encoder stem $E_\theta^{0:s}$, shared window $F_\theta^{s:e}$, tail $E_\theta^{e:L}$, projector $P_\psi$, language model $D_\phi$, and LoopLook modules; maximum depth $R_{\max}$, exit threshold $\tau$, and $\epsilon>0$. Training additionally receives target response $\mathbf{Y}$ and optimization step $u$.}
\Output{Training loss $\mathcal{L}$, or visual tokens $\mathbf{V}$ and inference exit depth $\hat{R}$.}

\tcc{\textcolor{LLAccent}{Shared initialization}}
$\mathbf{q}\leftarrow\operatorname{TaskEncode}(\mathbf{x})$, \quad $\mathbf{H}^{0}\leftarrow E_\theta^{0:s}(\mathbf{I})$, \quad $K\leftarrow R_{\max}$\;

\tcc{\textcolor{LLAccent}{Task-conditioned recurrent perception}}
\For{$r=0,\ldots,R_{\max}-1$}{
$\mathbf{z}^{r}\leftarrow M_z([\mathbf{q};\operatorname{Pool}(\mathbf{H}^{r});\mathbf{e}_{r}])$\;
$(\gamma_{\ell}^{r},\beta_{\ell}^{r})\leftarrow M_{\ell}(\mathbf{z}^{r})$ for each block $\ell$ in the recurrent window\;
$\operatorname{AdaMod}_{\ell}^{r}(\mathbf{Z})\coloneqq(1+\gamma_{\ell}^{r})\odot\operatorname{LN}(\mathbf{Z})+\beta_{\ell}^{r}$ for each block $\ell$\;
$\mathbf{H}^{r+1}\leftarrow F_\theta^{s:e}(\mathbf{H}^{r};\{\operatorname{AdaMod}_{\ell}^{r}\}_{\ell})$\;
$s^{r+1}\leftarrow\sigma(h_\omega(\operatorname{Pool}(\mathbf{H}^{r+1}),\mathbf{q},\mathbf{e}_{r+1}))$\;
\If{$r+1=R_{\max}$}{
$s^{r+1}\leftarrow 1$\;
}
\If{$m=\mathrm{infer}$ \textbf{and} $(s^{r+1}>\tau\ \textbf{or}\ r+1=R_{\max})$}{
$K\leftarrow r+1$\;
\textbf{break}\;
}
}

\tcc{\textcolor{LLAccent}{Shared prior-aware Mixture-of-Looks readout}}
\If{$m=\mathrm{infer}$}{
$\hat{R}\leftarrow K$\;
}
$\pi_r\leftarrow s^{r}\prod_{j<r}(1-s^{j})$ for $r=1,\ldots,K$\;
\For{$i=1,\ldots,N$}{
$\alpha_{i,r}\leftarrow\operatorname{softmax}_{r=1,\ldots,K}\!\left(g_\eta(\mathbf{h}_{i}^{r},\mathbf{q},\mathbf{p}_{i},r)+\log(\pi_r+\epsilon)\right)$\;
}
$\tilde{\mathbf{h}}_i\leftarrow\sum_{r=1}^{K}\alpha_{i,r}\mathbf{h}_{i}^{r}$ for $i=1,\ldots,N$, \quad $\tilde{\mathbf{H}}\leftarrow\{\tilde{\mathbf{h}}_i\}_{i=1}^{N}$\;
$\mathbf{V}\leftarrow P_\psi(E_\theta^{e:L}(\tilde{\mathbf{H}}))$\;

\tcc{\textcolor{LLAccent}{Training objective or standard language generation}}
\uIf{$m=\mathrm{train}$}{
$\lambda_d(u)\leftarrow\lambda_d^{\star}\min\!\left(1,\max\!\left(0,\frac{u-t_w}{t_r}\right)\right)$\;
$\mathcal{L}_{\mathrm{NTP}}\leftarrow-\sum_{t=1}^{T_y}\log p_\phi(y_t\mid y_{<t},\mathbf{x},\mathbf{V})$\;
$\mathcal{L}\leftarrow\mathcal{L}_{\mathrm{NTP}}+\lambda_d(u)\sum_{r=1}^{R_{\max}}r\,\pi_r$\;
\Return{$\mathcal{L}$}\;
}
\Else{
\Return{$(\mathbf{V},\hat{R})$ for generation with $D_\phi(\cdot\mid\mathbf{x},\mathbf{V})$}\;
}
\end{algorithm*}

\begin{table*}[t]
    \centering
    \caption{Benchmarks used in our evaluation.}
    \label{tab:app_benchmarks}
    \small
    \setlength{\tabcolsep}{12pt}
    \renewcommand{\arraystretch}{1.25}
    \begin{tabular*}{0.92\textwidth}
        {@{\extracolsep{\fill}} l l r l @{}}
        \toprule
        \textbf{Benchmark} & \textbf{Group} &
        \textbf{\# Samples} & \textbf{Format} \\
        \midrule
        MMStar         & General                  & 1,500 & MC (letter) \\
        VMCBench (DEV) & General                  & 1,000 & MC (letter) \\
        AI2D           & General                  & 3,088 & MC (letter) \\
        V$^\ast$       & Perception and Reasoning &   191 & MC (letter) \\
        BabyVision     & Perception and Reasoning &   388 & MC (letter) / Open-ended \\
        VisualPuzzles  & Perception and Reasoning & 1,168 & MC (letter) \\
        POPE           & Hallucination            & 9,000 & Yes/no \\
        HallusionBench & Hallucination            &   951 & Yes/no \\
        \bottomrule
    \end{tabular*}
\end{table*}
\section{Baselines and Benchmark Details}
\label{sec:app_baselines}

\subsection{Baseline Methods}
\label{sec:app_baselines_methods}

We compare LoopLook with three visual-side enhancement methods:
Dense Connector \citep{yao2024dense}, TGIF \citep{lin2026text}, and DeepStack \citep{meng2024deepstack}. All three provide public
checkpoints built on the same Vicuna-7B-v1.5 \citep{touvron2023llama2openfoundation} language backbone, enabling comparison with existing approaches to multi-layer visual feature integration, instruction-guided layer fusion, and intermediate visual feature injection.
Their visual pathways differ: Dense Connector adopts a
SigLIP-SO400M tower, and DeepStack uses the LLaVA-NeXT \citep{liu2024llavanext} recipe
with AnyRes encoding.

\paragraph{Dense Connector.}
We use the released \texttt{DenseConnector-v1.5-7B} checkpoint. The original LLaVA-1.5-7B uses CLIP ViT-L/14-336 \citep{radford2021clip}; this variant employs SigLIP-SO400M \citep{zhai2023sigmoid} at $384^2$ resolution. It aggregates features from three groups of encoder depths and concatenates them channel-wise before the MLP projector. The resulting representation has $3\times1152=3456$ dimensions per patch, producing $729$ visual tokens per image. The fusion is independent of the input instruction, and the visual token count is fixed.

\paragraph{TGIF.}
TGIF (Text-Guided Layer Fusion) uses the input instruction to determine the contribution of each vision encoder layer. A two-layer MLP maps the mean-pooled prompt embeddings to $24$ layer weights, which are normalized by softmax. The weighted layer features are combined with the penultimate-layer features and passed through the standard MLP projector. TGIF retains the CLIP ViT-L/14-336 backbone and produces $576$ visual tokens, without changing the encoder depth or token count at inference time.

\paragraph{DeepStack.}
We use the released high-resolution checkpoint \texttt{deepstack-l-hd-vicuna-7b}. DeepStack employs AnyRes-style multi-crop encoding and injects additional visual features into the early layers of the language model through residual connections. Its visual sequence includes a global image view and local crops, and auxiliary features provide additional visual detail at multiple scales. DeepStack therefore also changes how visual information enters the language model. Its visual token count varies with input resolution.

\subsection{Benchmark Details}
\label{sec:app_baselines_benchmarks}

We evaluate all methods on eight benchmarks covering general visual understanding, perception and reasoning, and hallucination robustness. We use the same prompts and answer extraction rules across methods. We follow the prescribed answer format for each benchmark. Multiple-choice questions are evaluated by option accuracy, while open-ended questions are evaluated against their reference answers. POPE and HallusionBench use binary yes/no answers.

\paragraph{General understanding.}
\textbf{MMStar} \citep{chen2024we} comprises $1{,}500$ multiple-choice questions across diverse visual understanding categories, including visual reasoning, mathematics, and knowledge-intensive tasks. Its construction filters out questions that can be answered from text alone, making it useful for assessing whether models can effectively ground their answers in visual evidence. \textbf{VMCBench} \citep{zhang2025automatedgenerationchallengingmultiplechoice} consolidates existing visual question answering datasets into a unified multiple-choice format, covering diverse sources such as charts, documents, OCR, and natural images. We use its DEV split of $1{,}000$ samples to evaluate performance across heterogeneous visual question answering tasks. \textbf{AI2D} \citep{10.1007/978-3-319-46493-0_15} focuses on grade-school science diagrams, where answering questions requires interpreting diagram structure, textual annotations, and spatial relations indicated by visual elements such as arrows. It complements natural-image benchmarks by testing the understanding of structured visual representations.

\paragraph{Perception and Reasoning.}
\textbf{V$^\ast$} \citep{wu2024v} evaluates fine-grained visual perception through questions about local attributes, relative positions, text within images, and challenging examples drawn from the GPT-4V evaluation set. These questions require models to identify specific visual evidence beyond global scene understanding. \textbf{BabyVision} \citep{chen2026babyvision} targets basic visual abilities through questions designed to minimize dependence on external knowledge, covering elementary forms of visual recognition and reasoning. Its tasks help assess whether models can reliably interpret simple visual evidence, even when they perform well on broader vision-language benchmarks. \textbf{VisualPuzzles} \citep{song2025visualpuzzlesdecouplingmultimodalreasoning} consists of non-verbal visual puzzles that emphasize pattern induction, spatial composition, and counting. By reducing reliance on linguistic context and specialized knowledge, it probes the ability to infer visual rules and reason over image structure.

\paragraph{Hallucination robustness.}
\textbf{POPE} \citep{li2023evaluating} evaluates object hallucination by asking binary questions about whether specified objects are present in COCO \citep{lin2014microsoft} images. Its random, popular, and adversarial sampling strategies vary the difficulty of negative examples, testing whether a model distinguishes actual visual evidence from object co-occurrence priors. \textbf{HallusionBench} \citep{guan2024hallusionbench} examines whether models are misled by visual illusions, misleading image-text associations, and conflicts between visual evidence and language priors. Its questions cover cases in which intuitive or language-driven answers can contradict what the image actually depicts.

\subsection{Benchmark Prompts}
\label{sec:app_bench_prompts}

All eight benchmarks follow the prompt templates released in VLMEvalKit, and we adopt them verbatim in every evaluated setting. Here we list each template: \texttt{\{question\}} and \texttt{\{options\}} denote the per-sample question and choice list, and the image is prepended to the text prompt. Yes/no benchmarks (POPE, HallusionBench) use the annotation question as released by VLMEvalKit without any additional suffix.

\begin{tcolorbox}[title={MMStar / AI2D / V$^\ast$ (multiple-choice template)}, colbacktitle=LLMain, fonttitle=\small\bfseries]
\begin{promptbox}[Template]
Question: \{question\}\\
Options:\\
A. \{option\_A\}\\
B. \{option\_B\}\\
...\\
Please select the correct answer from the options above.
\end{promptbox}
\end{tcolorbox}

\begin{tcolorbox}[title={VMCBench (multiple-choice template)}, colbacktitle=LLMain, fonttitle=\small\bfseries]
\begin{promptbox}[Template]
Question: \{question\}\\
Options:\\
A. \{option\_A\}\\
B. \{option\_B\}\\
...\\
Answer with the option's letter from the given choices directly.
\end{promptbox}
\end{tcolorbox}

\begin{tcolorbox}[title={VisualPuzzles}, colbacktitle=LLMain, fonttitle=\small\bfseries]
\begin{promptbox}[Template]
\{question\}\\
Options:\\
(A) \{option\_A\}\\
(B) \{option\_B\}\\
(C) \{option\_C\}\\
(D) \{option\_D\}\\
Solve the multiple-choice question and then answer with the option letter from the given choices. The last line of your response should be of the following format: `Answer: \$LETTER' (without quotes) where LETTER is one of the options. Think step by step before answering.
\end{promptbox}
\end{tcolorbox}

\begin{tcolorbox}[title={POPE / HallusionBench (yes/no)}, colbacktitle=LLMain, fonttitle=\small\bfseries]
\begin{promptbox}[Template]
\{annotation question\}\\
\emph{e.g., POPE: Is there a \{object\} in the image?}
\end{promptbox}
\end{tcolorbox}

\begin{tcolorbox}[title={BabyVision}, colbacktitle=LLMain, fonttitle=\small\bfseries]
\begin{promptbox}[Template]
\{question\}\ \ (\emph{open-ended items})\\[2pt]
\{question\}\\
Choices:\\
(A) \{option\_A\}\\
(B) \{option\_B\}\\
...\\
Think about the question and give your final answer in \textbackslash boxed\{Answer\} format.
\end{promptbox}
\end{tcolorbox}

\end{document}